\documentclass{article}

\usepackage[english]{babel}

\usepackage[letterpaper,top=2cm,bottom=2cm,left=3cm,right=3cm,marginparwidth=1.75cm]{geometry}

\usepackage{amsmath}
\usepackage{graphicx}
\usepackage{hyperref}
\hypersetup{colorlinks,allcolors=blue}

\usepackage[square,numbers]{natbib}
\usepackage{authblk}

\usepackage{float}
\usepackage{amsmath}
\usepackage{graphicx}
\usepackage{xcolor}
\usepackage{multirow}
\usepackage{multicol}
\usepackage{caption}
\usepackage{subcaption}
\usepackage{tikz}
\usepackage{array}
\usepackage{appendix}
\usepackage{etoolbox}

\gappto{\UrlBreaks}{\UrlOrds}

\newcommand{\quotes}[1]{``#1''}

\title{LLMs Perform Poorly at Concept Extraction in Cyber-security Research Literature}

\author[1]{Maxime Würsch}
\author[1, 2, +]{Andrei Kucharavy}
\author[1,2]{Dimitri Percia David}
\author[1]{Alain Mermoud}

\affil[+]{Corresponding Author; andrei.kucharavy@hevs.ch}
\affil[1]{Cyber-Defence Campus, armasuisse S+T}
\affil[2]{Institute of Entrepreneurship \& Management, HES-SO Valais-Wallis}
\affil[3]{Section of Computer Science, EPFL}

\begin{document}
\maketitle

\begin{abstract}
\noindent
The cybersecurity landscape evolves rapidly and poses threats to organizations. To enhance resilience, one needs to track the latest developments and trends in the domain. It has been demonstrated that standard bibliometrics approaches show their limits in such a fast-evolving domain. For this purpose, we use large language models (LLMs) to extract relevant knowledge entities from cybersecurity-related texts. We use a subset of arXiv preprints on cybersecurity as our data and compare different LLMs in terms of entity recognition (ER) and relevance. The results suggest that LLMs do not produce good knowledge entities that reflect the cybersecurity context, but our results show some potential for noun extractors. For this reason, we developed a noun extractor boosted with some statistical analysis to extract specific and relevant compound nouns from the domain. Later, we tested our model to identify trends in the LLM domain. We observe some limitations, but it offers promising results to monitor the evolution of emergent trends.
\end{abstract}

\section{Introduction}
\subsection{Bibliometrics-based technological forecasting}
Secure and reliable information systems have become a central requirement for the operational continuity of the vast majority of goods and services providers \cite{percia2020three}. However, securing information systems in a fast-paced ecosystem of technological changes and innovations is hard \cite{anderson_security_2020}. New technologies in cybersecurity have short life cycles and constantly evolve \cite{daim_digital_2022}. This exposes information systems to attacks that exploit vulnerabilities and security gaps \cite{anderson_security_2020}. Hence, cybersecurity practitioners and researchers need to stay updated on the latest developments and trends to prevent incidents and increase resilience \cite{daim_anticipating_2016}.

A common approach to gather cured and synthesized information about such developments is to apply bibliometrics-based knowledge entity extraction and comparison through embedding similarity \cite{chen2021topic, safder2019bibliometric, zhang2018does} --  recently boosted by the availability of entity extractors based on large language models (LLMs) \cite{dunn2022structured, petroni2019language}. However, it is unclear how appropriate this approach is for the cybersecurity literature. We address this by emulating such an entity extraction and comparison pipeline, and by using a variety of common entity extractors -- LLM-based and not --, and evaluating how relevant embeddings of extracted entities are to document understanding tasks -- namely classification of \textit{arXiv} documents as relevant to cybersecurity (\url{https://arxiv.org}).

While LLMs burst into public attention in late 2022 -- in large part thanks to public trials of conversationally fine-tuned LLMs \cite{InstructGPT2022OpenAI, AIBorder2022Anthropic, LIAON2023OpenAssistantConvos}--, modern large language models pre-trained on large amounts of data trace their roots back to ELMo LLM, first released in 2018 \cite{ElMo2018}. Used to designate language models with more than approximately 100M parameters and pretrained on 1B or more tokens \cite{StochasticParrots2021BenderGebruMitchell, DistilBERT2019} -- such as BERT or RoBERTa \cite{BERT2018, RoBERTa2019}--, smaller LLMs have proven to provide a valuable insight into the behavior and capabilities of larger ones \cite{ScalingLawsLLM2020OpenAI, ComputeOptimalLLMs2022Google, GPT42023OpenAI}, presenting both a weaker version of larger model capabilities, but also more mild version of larger model failure modes \cite{UnpredictabilityOfLLMs2022Anthropic}. In this paper, we focus on smaller LLMs, ranging from 110M to 350M parameters trained for a variety of task extractions, ranging from noun and keyphrase extraction to named entities, concept recognition, to token classification. To provide a comparison reference, we also included two non-LLM-based approaches, namely Yake keyphrase extractor \cite{yake2020} and SpaCy noun extractor \cite{spaCy2020}. Not only does this approach allow us to glean insight into larger LLMs, but it is also a current standard and allows a more scalable and less resource-intensive entity extraction, making it particularly appealing.

Unfortunately, while widely adopted, this approach does not work well for scientific bibliometrics. We show that despite the apparent abundance of available models, LLM-based entity extractors are extremely similar for similar tasks due to base models and fine-tuning datasets re-use. We then show that such models are ill-suited for bibliometrics tasks not only in cybersecurity-related topics but in computer science research in general, in part due to the nature of those fine-tuning datasets, calling into question this approach. We then show that even if we assume the relevance of extracted terms, their downstream automated processing remains a challenging task, given that it is highly sensitive to the embedding choice. Overall, we believe our results call into question the usage of LLMs for entity extraction in scientific literature.

To this end, we test a bibliometrics-based approach to extract controlled vocabulary terms from scientific texts on cybersecurity using large language models (LLMs). Specifically, we extract entities such as keywords, nouns, or named entities from the body of a publication with pre-trained models (such as with Yake or KeyBERT \cite{yake2020, grootendorst2020keybert}); we apply embedding of extracted keywords into a vector space (with algorithms such as word2vec \cite{Word2Vec2013}) to allow for a more straightforward comparison of academic publications and entities cited in them; and we use unsupervised learning of entity relations through means such as aggregating low-dimensional projection (such as t-SNE or UMAP \cite{TSNE2008, UMAP2018}). We aim to identify and track cybersecurity concepts and topics over time by extracting keywords from texts that represent technologies, entities, and interactions. We use a subset of arXiv preprints (100k) on cybersecurity as our text source. We compare different LLMs on named entity recognition (NER) and relevance. NER identifies and classifies entities from texts, while relevance measures how well the keywords match the texts’ theme and purpose \cite{li2020survey}.

Our results show that LLMs do not produce relevant keywords that reflect the cybersecurity technological context. This is the reason that in the second part, we explore the usage of noun extraction in more detail, as they look promising, as discovered in the first part. We have developed a model with the help of spaCy \cite{spaCy2020} that extracts compound nouns, such as \quotes{high school}. We later filter the extracted entities by comparing their frequency against the BookCorpus using a volcano plot to obtain only the terms relevant to the cybersecurity domain. We evaluate our model by analyzing the evolution of technologies pivotal to the emergence of the LLMs. We observe that the model still has some flaws, but the results look promising to create an efficient tool to forecast emerging trends in a fast-evolving domain like cybersecurity. We later discuss the reasons for this limitation and suggest future research directions.

\section{Background} \label{background}

Usually, standard methods for forecasting and monitoring trends of different technologies use data from platforms such as OpenAlex and Google Trends. Those proxies have shown in the past to be effective instruments to decipher the evolution's trends of different technologies~\cite{daim_digital_2022, adner2002emergence, klepper1997industry, perez2010technological, rogers_diffusion_2010}. Measuring the adoption and development of the technologies that are pivotal to the evolution of LLMs should be no different~\cite{chumnumpan2019understanding, jun2018ten, woloszko2020tracking, OpenAlex2023, priem2022openalex}. However, those methods have been shown to provide inadequate trends and have difficulty deciphering the rapid development in the LLM domain. Recent studies demonstrate that the trends provided by proxies measuring attention, such as citation counts, suffer from delay and sparse data~\cite{kucharavy_fundamentals_2023}. We performed an analysis of the performance and effectiveness of Google Trends and OpenAlex on keywords linked to technologies that participate in the evolution of the LLM, according to experts in the LLM domain. As Google Trends uses keywords directly, we just need to pass the names of the technologies under watch. Since OpenAlex works with a topic ontology in the background, some work is required to align the keyword to the ontology. The keyword under watch: \textit{Neural Language Model}, \textit{Deep Neural Language Model}, \textit{Attention}, \textit{Self-Attention}, \textit{Transformer Model}, \textit{Large Language Model}, \textit{Fine-Tuning}, \textit{Transfer Learning}, and \textit{Conversational Agent}. The resulting trends of our analysis are set to be compared to those recognized by experts in the LLM domain.

\subsection{Google Trends}

\begin{figure}[h]
\centering
\textbf{Google Trends for specified terms}\par\medskip
\includegraphics[width=0.95\textwidth]{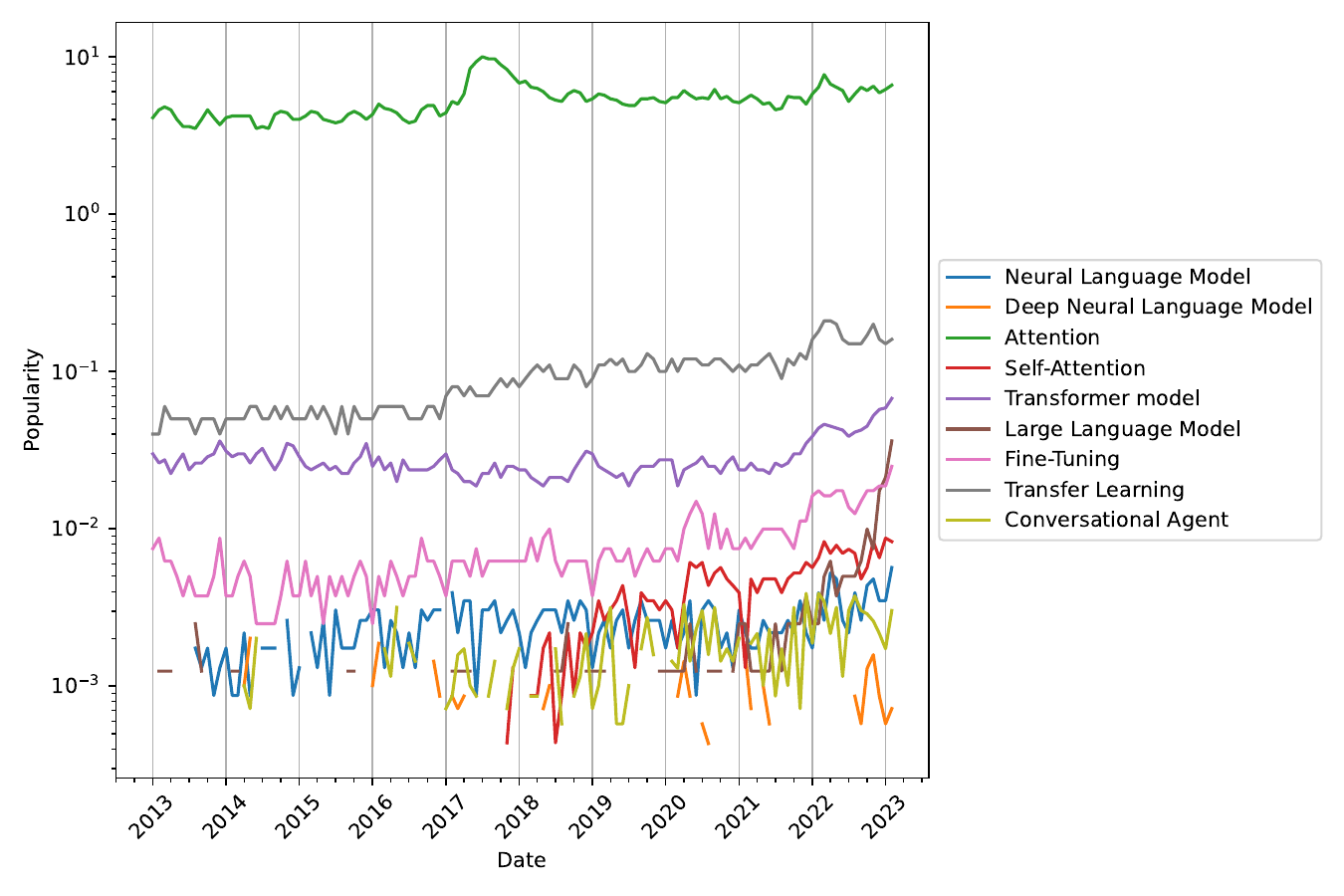}
\caption{\label{fig:proxy_google_trends} Attention capture by Google Trends, 2013--2023}
\end{figure}

Google Trends provides trends that follow the evolution identified by experts. \textit{Attention}, \textit{Transfer Learning}, and \textit{Transformer Model} reveals high level of noise due to semantic contamination. Nevertheless, some trends using specific terminologies match the trends identified by the experts. However, direct utilization of Google Trends is limited in the precision of the results, as they are relative and rounded to the nearest integer. The quantity of analyzed keywords is also limited. We bypass these limitations by utilizing the \emph{g-tab} tool~\cite{gtab}. Unfortunately, some challenges remain, such as the temporal lag for public attention to discover the trends and the impossibility of specifying the domain of the keywords, leading to the contamination of some terms.

\subsection{OpenAlex}

\begin{figure}[ht]
\centering
\textbf{Normalized OpenAlex citation trends for citation of specified terms}\par\medskip
\includegraphics[width=0.95\textwidth]{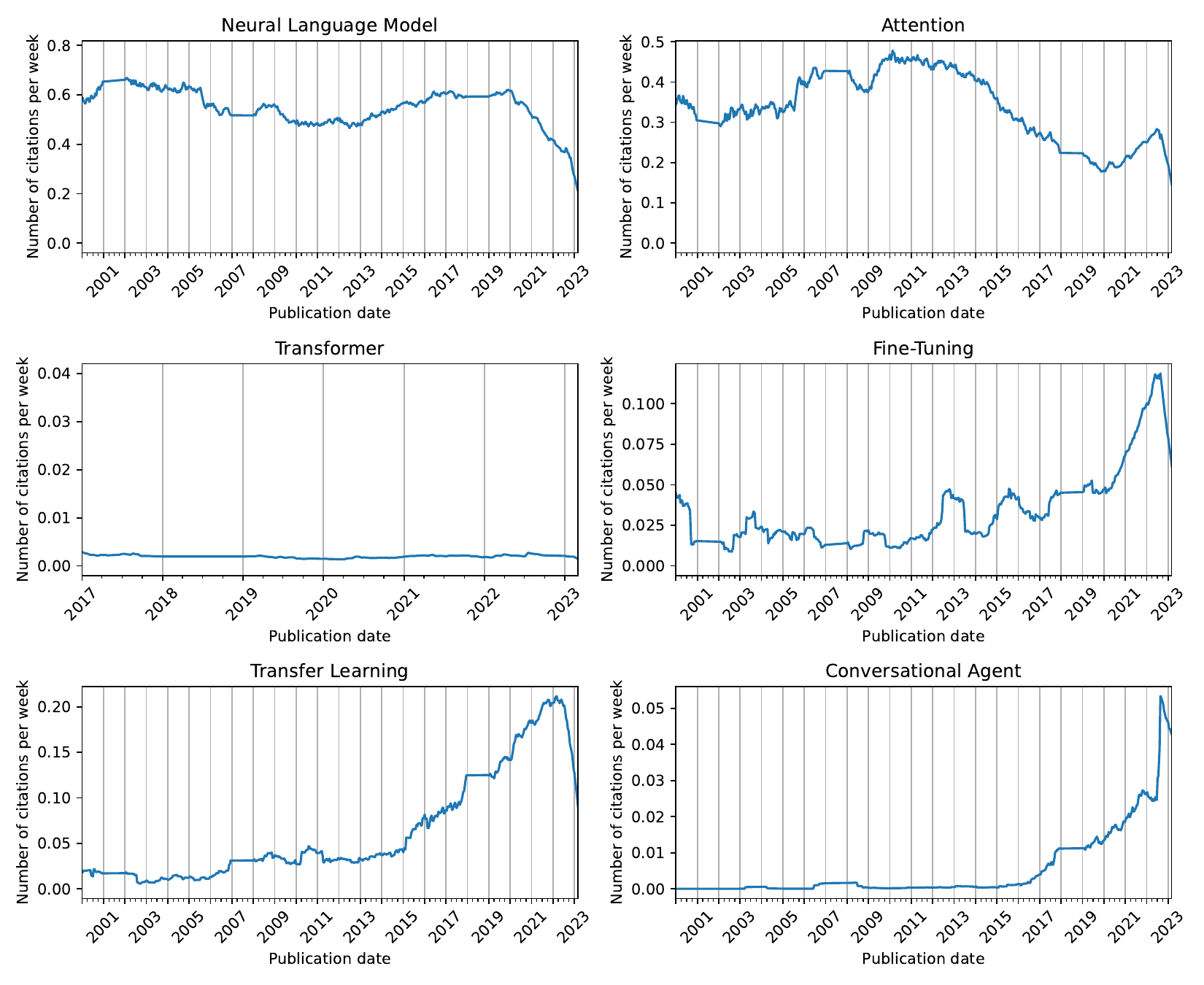}
\caption{\label{fig:proxy_openalex} The citation count for academic papers on OpenAlex, segmented by the week of publication, filtered by their ontology correlated to the provided keywords. The timeframe is from 2000 until now except the \emph{Transformer} one, which is limited to 2017-now to mitigate the contamination of unrelated publications.
}
\end{figure}

As opposed to Google Trends, OpenAlex requires to match the keywords on its topic ontology. Unfortunately, the insufficient granularity and the difficulty in matching keywords to categories due to descriptions that are too short represent a challenge to measuring the trends, therefore requiring manual inspection. Inherent delays and periodic variation induced, for example, by conference cycles, require rigorous post-processing, such as normalization and seasonal decomposition, to mitigate their impact. Even with these efforts, analyzing trends in the last three years is complex; some categories, such as \emph{Transformer}, are too contaminated to provide valuable results.

\subsection{Summary and Discussion}

The limitation observed during the analysis of trends on the evolution of the LLMs illustrates the limitation of conventional trend analysis methods, resulting in the need to develop new tools to accurately forecast emerging trends in LLMs. Specifically, the following requirements are important to be used in developing the new methods. We believe that observed issues happen in any domain with a short cycle of development.

\begin{itemize}
    \item Due to the delay for a new technology to emerge and the time for it to get a consistent name, it should not depend on a single ontology
    \item Considerate the delay for data recording to prevent biases against recent results
    \item Take into account for the semantic drift, such as \quotes{Transformer-Like} and \quotes{Self-Attention} are correctly designate \quotes{Neural Language Models}
    \item Topics need to be accurately separated. For example, Transformer Neural Language Models need to be isolated from current Transformers or other types of Transformers from unrelated fields.
\end{itemize}

\section{Literature Review}

\subsection{Bibliometrics and Public Attention Measurement}

There is ongoing research in the bibliometrics domain. Multiple papers have been published in recent years evaluating new techniques or enhancing already broadly used techniques. The work of Donthu et al., that have been published in 2021~\cite{donthu_how_2021}, explore and provide a methodology on how to use the emergent wide variety of database and the new proliferating bibliometrics software. On the other side, Daim and Yalçin explain in 2021 that there is no real border to the bibliometrics domain~\cite{daim_digital_2022}. Other domains have different names but have the same aims as bibliometrics using different types of metrics, such as \emph{altmetrics} that perform similar analysis using less recognized sources of information such as social media or GitHub's repositories. They also demonstrate an efficient method that scrutinizes networks to augment the quality of bibliometrics analysis.

Jun et al. evaluate the usage of the well-known Google Trends website over ten years~\cite{jun_ten_2018}. In particular, its ability to analyze big data. They came to the conclusion that Google Trends is an efficient tool for trend monitoring. However, it is crucial to keep in mind its limitations. Namely, the emotional impact of the trend. For example, trends linked to a political election are linked to the emotion at the instant of the request and not the real vote intention of the Google search engine user.

\subsection{Entity Extractors Evaluation}

It has been demonstrated that it is hard to compare the difference and quality of entity extractors using well-used benchmark datasets~\cite{sun_review_2020}. This issue is due to the fact that the extractor extracts semantically similar sentences most of the time but not the one present in the evaluation dataset. They came to this conclusion after comparing a large variety of non-LLM-based entity extractors. They also found that the overall structure and the length of the document, particularly when it contains multiple topics, can significantly decrease the performance of the extractor. Svrvastava et al. demonstrate the importance of using specialized fine-tuned datasets when training the extractor~\cite{srivastava_study_2023}. The precision of the model can be augmented between 2\% and 7\% by using those datasets. Therefore, it illustrates the difficulty of discovering new trends. It is crucial to get the best performance to regularly re-fine-tune the model on the latest data (which is applicable to LLMs in general and is implemented in practice). Unfortunately, it is time-consuming and costly to keep the model and the fine-tuning dataset. Scientists have organized a workshop to discuss the challenge to resolve in order to obtain better information retrieval (IR) and the implication of LLMs in the domain~\cite{ai_information_2023}. Among the resulting challenges to solve, they observe that LLMs have difficulty extracting high-quality entities, especially in specific domains. They also found some similar issues as me for the performance of LLM-based entity extractors.

\subsection{Alternative Extractors}

A NER extractor specialized in the cyber security domain has been developed by Gao and Zhang using the long short-term memory (LSTM) model, the ancestor of the transformer architecture used in LLMs~\cite{gao_data_2021, vaswani_attention_2017}. Their model, using a dictionary feature embedding and a multi-head attention window, is able to, with decent precision, the seven different entity types defined in their dataset: application, software version, hardware type, operating system, edition file type, and vendor. Despite its quality, the model has two notable downsides. One is its difficulty in recognizing entities that do not often appear on the training set. The second one is that the problem cannot be generalized to extract entities it has not been trained on. This implies the need for a large supervised dataset, which is regularly updated to keep track of the new technologies. Maintaining such a dataset is complex and a bit against what we try to achieve. We want to detect emerging technologies. Therefore, we cannot depend on an annotated dataset to discover the emergent technologies we do not know. The model they present still has some helpful usage, such as for analyzing firms' attack software. Since most of the time, the attacker uses well-known vulnerabilities that are stored in the CVE database. It is a database of known vulnerabilities in standard software. One of the data sources of the dataset used to train the model.

Old techniques developed before the LLM era still provide decent results, such as the latent Dirichlet allocation (LDA) for topic modeling~\cite{blei_latent_2003}. This method has been used to scrutinize the impact of the COVID-19 pandemic on the cybersecurity domain by comparing the state before and after the pandemic. LDA extracts the proportion of $k$ topic inside each provided document. The set of topics is the same for all the documents. The number $k$ needs to be selected wisely to obtain the best performance as it is a hyper-parameter of the algorithm. For specific trend monitoring, the number $k$ will need to be large to get more precise topics. Unfortunately, it augments the probability of the model to over-fit the data. Hence, this technique does not seem well suited for monitoring emerging trends.

\section{Methods}

The complete code to replicate the results presented here with instructions is available at \url{https://anonymous.4open.science/r/joi-28CE/}.

\subsection{Dataset}
The dataset used is a copy of arXiv preprints until September 2021, initially collected by \citep{david2023measuring}. We focused on the \textit{cs} category, specifically on the cs.CR and cs.NI listings - Cryptography and Security and Network and Internet Architecture, as most relevant to cybersecurity. In addition to them, we added six additional unrelated listings (cs.CC, cs.LO, cs.DS, cs.IT, cs.CL, and cs.AI) as comparison domains. The selected listings represented $5000$ to $20000$ preprints each.

\subsection{Processing of the input text}

The data of arXiv are in PDF format. Therefore, we extract the text by using \emph{fitz}, a Python module for PDF file manipulation. Afterward, we used an LLM model trained for language recognition, based on XLM-RoBERTa fine-tuned on a dataset for language identification, available at \url{https://huggingface.co/papluca/xlm-roberta-base-language-detection}, to keep only English text. The preamble of the preprint and the references are removed to get the best performance of the keyword extractor. For this, a simple word search has been used. Since the text extraction from PDF is not perfectly efficient, we need to do some post-processing to remove unwanted characters or missed transformations to text. Those issues arrive mainly in the tables and the mathematical formula. Another issue is that the extracted text contains many returns lines in the middle of sentences due to the wrapping to stay in the page frame. It is crucial to remove them to reduce the error of the extractor. A simple word replacement has been used.

\subsection{Entity Extraction}

Following that, we applied models described in \autoref{table:model_refs} to the documents. Specifically, four major classes of models were used: Noun Extractrs (NnE), Keyphrase Extractors (KPE), Named Entity Recognition (NER), and Token Classification (TokC). Two NER models performed additional tasks: number recognition (NER + NUM) and concept recognition (NER + CON R). Those different types of entity extractors are explained below. Their principal aim is to analyze a provided text input to extract relevant information. The type of the extracted tokens depends on the model category. The full description of extractors used is provided in \autoref{table:model_refs}.

\begin{description}
\item[Noun Extractor (NnE):] Extract the noun of a sentence by analyzing its structure. With their analysis, it is also possible to extract compound nouns such as High school.
\item[Keyphrase Extractor (KPE):] Extract some sentences from the text to summarize the document totally in the best possible way.
\item[Token Classification (TokC):] Extract and then classify the tokens by assigning them labels.
\item[Named Entity Recognition (NER):] It is similar to token classification as it is a subtask. The model extracts entities corresponding to categories such as firms, date, person names, numbers, \ldots, on which the model has been trained on.
\item[Concept Recognition (CON + R):] Retrieve the different concepts that are present in the provided text.
\end{description}

In addition to the type of extractors, we can divide them into three different categories. First, the ones that come from HuggingFace. A website containing multiple pre-trained LLMs. We used name entity recognition (NER) to get the keywords. Only the terms with the best score were kept. The second category is the keywords extractor model. There are two of them: KeyBert, which is a model based on a HuggingFace model but adds some functionalities, and Yake, which is one of the state-of-the-art unsupervised non-LLM keyword extractors. The last category is noun extractor. We extracted the nouns that appeared the most in the document as keywords. spaCy is the model used for this.

For LLM-based entity recognition models, the documents were segmented to allow the text to fit the attention window of the LLM model fully. If the number of entities extracted from the document exceeded 100, only 100 entities with the highest confidence scores were retained. Confidence scores were derived as LLM model final layer activation before the finalization step. Samples of extracted entities are available in the sample file in the code repository.

\begin{table}[h!]
\begin{center}
\begin{tabular}{||l | r | r | l||} 
 \hline
 Model Name & Refs & Entities/Doc & Type \\ [0.5ex] 
 \hline\hline
 {spaCy Large*$^p$} & \cite{spaCy2020} &  {$99.3 \pm 6.93$}  & \multirow{2}{4em}{Noun Extractor}\\ 
 {spaCy Transformer$^p$} & \cite{spaCy2020} &  {$99.3 \pm 6.97$}  & \\
 \hline
 {Yake*$^p$} & \cite{yake2020}&  {$19.9 \pm 1.97$} & \multirow{5}{4em}{Keyphrase Extractor}\\ 
 {KeyBERT$^p$} &  \cite{grootendorst2020keybert} &  {$99.3 \pm 7.25$}  & \\
 {KBIR kpcrowd} &  \cite{kbir22, KPCrowd2012} &  {$96.9 \pm 14.6$} & \\
 {KBIR inspec} &  \cite{kbir22, INSPECBis2019} &  {$76.4 \pm 27.7$} & \\
 {BERT-base-uncased} & \cite{BERT2018} &  {$44.7 \pm 24.0$} & \\
 \hline
 {BERT-base-uncased} & \cite{BERT2018} &  {$43.3 \pm 23.3$} & {NER+CON R} \\
 {XLM-RoBERTa-base Onconotes 5} & \cite{XLMRoBERTaBis2021, Ontonotes2006} &  {$36.4 \pm 23.4$} & {NER+NUM} \\ 
 \hline
 {ELECTRA-base conll03}  & \cite{Electra2020, Conll2003}&  {$39.9 \pm 25.0$} & \multirow{6}{4em}{NER}\\
 {BERT-large-cased conll03}  & \cite{BERT2018, Conll2003} &  {$41.7 \pm 24.9$} & \\
 {BERT-large-uncased conll03} & \cite{BERT2018, Conll2003} &  {$33.5 \pm 23.3$} & \\
 {DistilBERT-base-uncased conll03}  & \cite{DistilBERT2019, Conll2003} &  {$37.7 \pm 24.8$} & \\
 {RoBERTa-large conll03}  & \cite{RoBERTa2019, Conll2003} &  {$28.7 \pm 21.1$} & \\
 {XLM-RoBERTa-large conll03}  & \cite{XLMRoBERTaXL2021, Conll2003} &  {$26.0 \pm 19.5$} & \\
  \hline
 {BERT COCA-docusco} & \cite{BERT2018, docusco2012} &  {$99.6 \pm 6.11$} & {TokC}\\ [1ex] 
 \hline
\end{tabular}
\caption{Characterization of entity extractors analyzed. Models marked * are non-LLM based. Models marked $^p$ are part of Python repositories; all others were recovered as pre-trained weights from respective Huggingface repositories. Detailed repository links are available in the project code repository. Entities/Document is mean $\pm$ std.}
\label{table:model_refs}
\end{center}
\vspace*{-0.5cm}
\end{table}

\subsection{Noun Extraction}

\begin{figure}
\centering
\includegraphics[width=0.85\textwidth]{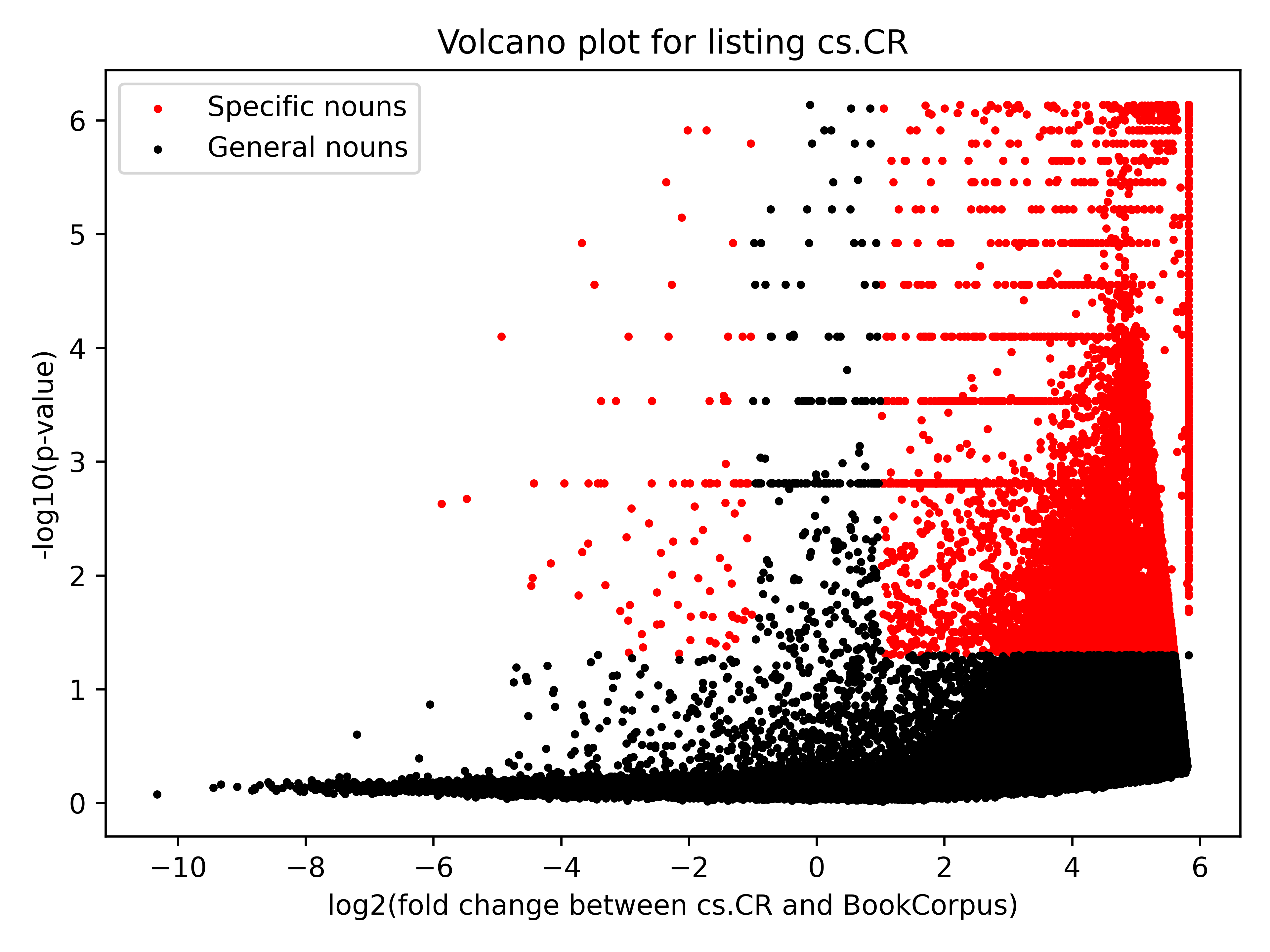}
\caption{2D projection with UMAP of spaCy embeddings of extracted entities.}
\label{fig:volcano_cr}
\end{figure}

In addition to testing existing entity extraction, we decided to create our own method. We use spaCy to extract the compound nouns, such as \quotes{high school}. It uses the structure and semantics of the sentence to do it. Extracting compound nouns allows us to get more context than just extracting nouns. To reduce the issues of PDF-to-text conversion, we require that the compound nouns appear at least three times in the corpus. We then applied a t-test and a fold change to measure the difference in frequency of the extracted keywords in arXiv against standard English text by using the BookCorpus. This corresponds to a volcano plot (cf. \autoref{fig:volcano_cr}). Only the words with a small p-value and a significant fold change are retained. The fold change corresponds to how the arXiv frequency differs from the BookCorpus frequency. The p-value is calculated in the following way. $X$ is a vector where each dimension represents the number of occurrences of the words in the different dataset. Then, a weighted mean and standard deviation are calculated. The weight corresponds to the number of words in each dataset. The value $n$ corresponds to minimal occurrences in the different datasets. The degree of freedom is n minus one, with a minimal value of zero. We repeat this calculation for each word present in the dataset.

$$\mathbf{t} = \frac{\mathbf{X} - x_{mean}}{x_{std} / \sqrt{n}}; \qquad\qquad df = max(0, n-1)$$

The p-value is finally calculated using the Student distribution's survival function scaled by the mean and standard deviation. We chose this distribution as it is the default one when we do not have values for the standard deviation.

\subsection{Comparison of Entity Extractors}

\begin{figure}[h]
\centering
\includegraphics[width=0.92\textwidth]{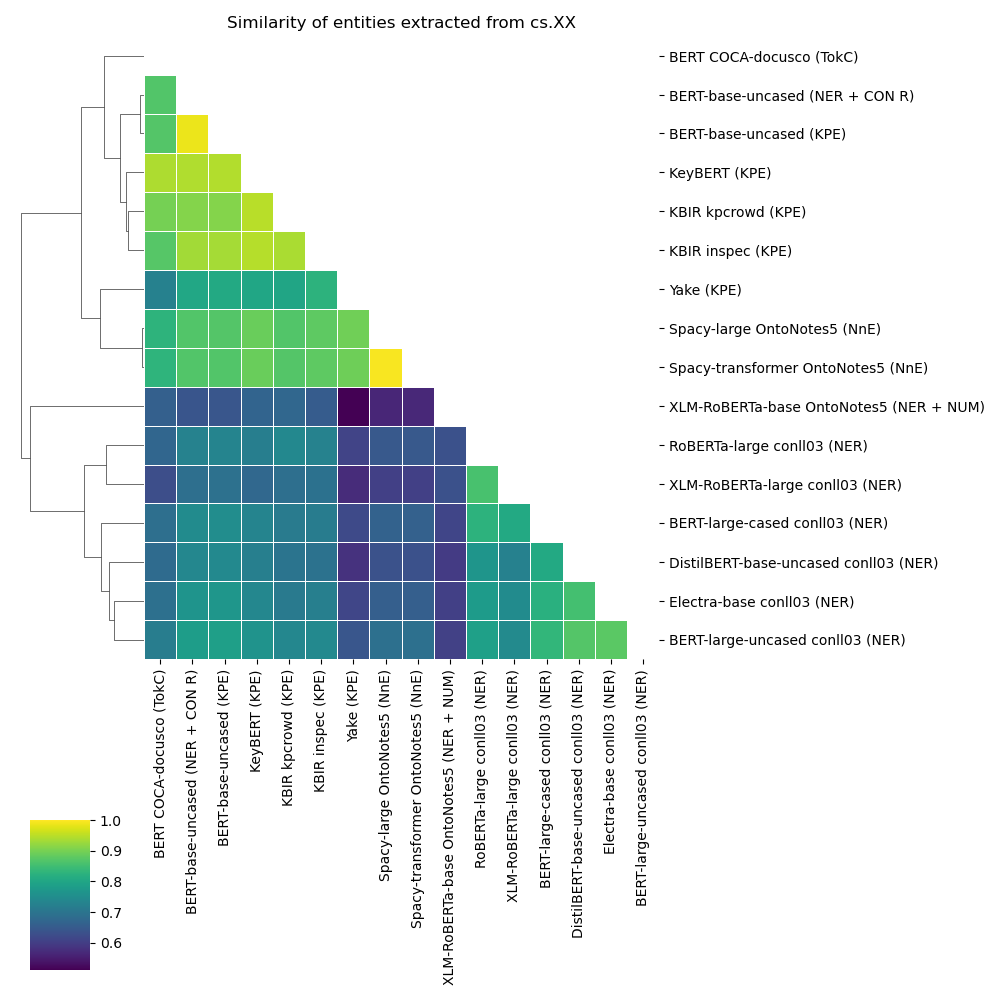}
\caption{Hierarchical clustering of entity extractors based on cosine similarity of terms they extract in spaCy embedding.}
\label{fig:extractor_clusters}
\vspace*{-0.25cm}
\end{figure}

We compare the performance of the entity extractors on all documents in the selected arXiv listings by embedding entities extracted from each document with spaCy and calculating the average cosine similarity between different extractors for each document. We then performed a hierarchical clustering on the average cosine similarity between each extractor using the single linkage algorithm (cf. \autoref{fig:extractor_clusters}). This is a good choice since it allows grouping the models by joining them by the most similar first. The average of all pair similarities is taken to compare two different models.

\subsection{Comparison of Extracted Entities}

To visualize the extracted entities, we used the common embeddings (spaCy \cite{spaCy2020}, GloVe  \cite{GloVe2014}, BERT-Large \cite{BERT2018}, GPT-2 \cite{GPT22019RadfordSutskeverOpenAI}, Fasttext \cite{FastText2017}, and word2vec \cite{Word2Vec2013}) and four low-dimensional projection algorithms (linear, spectral, t-SNE \cite{TSNE2008}, UMAP \cite{UMAP2018}) to investigate whether the entities extracted from preprints in different listings would correlate with the listings of preprints themselves. This approach allows us to evaluate whether the extraction and embedding would allow us to detect themes specific to different domains in an unsupervised manner. t-SNE and UMAP are highly sensitive controls, given that they are known to overfit underlying structures easily \cite{PachterEmbedding2021}. The 2D projection is, hence, a test as to whether an underlying conceptual structure is retained in a sufficient amount to allow for unsupervised paper theme similarity evaluation.

\begin{figure}
\centering
\includegraphics[width=0.85\textwidth]{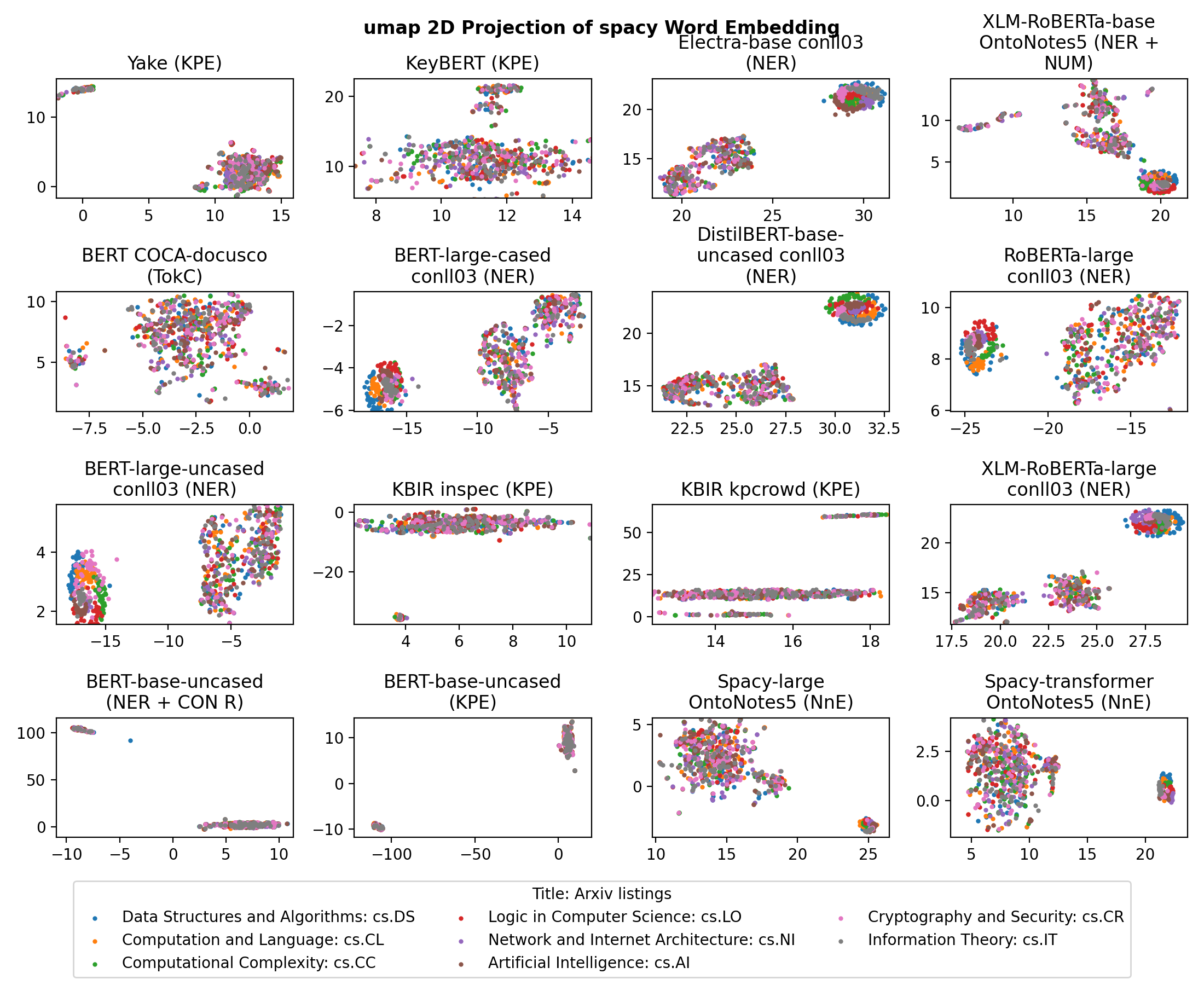}
\caption{2D projection with UMAP of spaCy embeddings of extracted entities.}
\label{fig:spacy_umap}
\end{figure}

\begin{figure}
\centering
\includegraphics[width=0.85\textwidth]{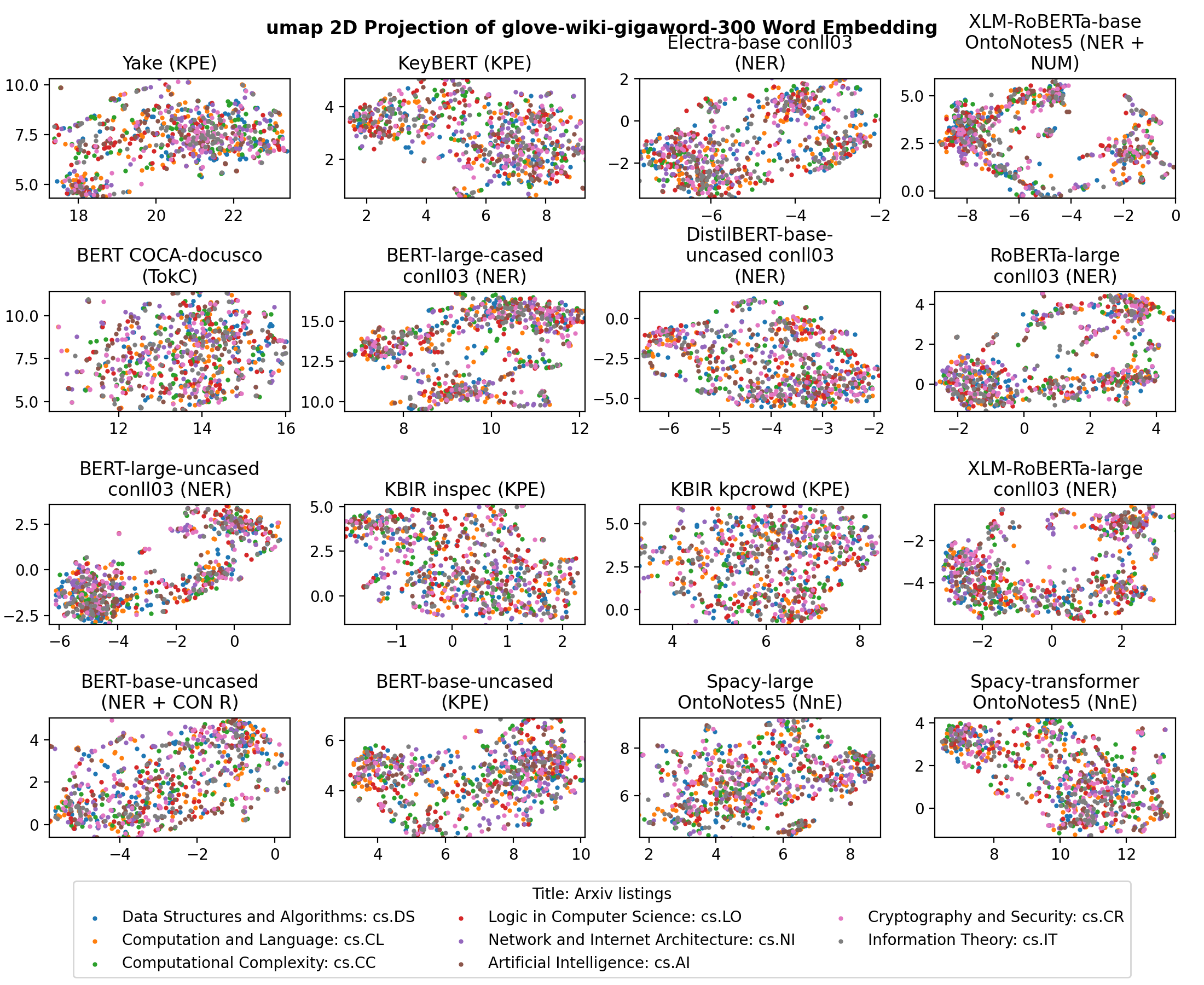}
\caption{2D projection with UMAP of GloVe embeddings of extracted entities.}
\label{fig:glove_umap}
\end{figure}

To allow the interpretation of the results, we subsampled 100 papers from each listing and, due to high processing time (cf. Figures \ref{fig:spacy_umap}, \ref{fig:glove_umap}; all figures in the code repository).

The data used in our pipeline to extract the keywords and draw plots come from the arXiv database \cite{david2023measuring}. We used a subset of about 10k preprints in the CS category. The pipeline can be divided into three parts. The first one processes the paper and extracts the keywords. Then, the correlation between the models is calculated and plotted. Finally, different embeddings are converted in a low-dimension space using different manifold algorithms.

\subsection{Comparison of the impact of the embedding}

To show the importance of using the same embedding as the one used for the keyword extraction, we load the keyword in different embedding and make a manifold clustering for each model. The points correspond to a keyword extracted from a preprint of arXiv, and the classes correspond to the listings of the cs arXiv category.

\subsection{Trends evolution of technologies using closeness in document}

To observe the development of various technologies close to the ones that lead to the emergence of the LLMs, mention in the background section (\autoref{background}). We examined the specific compound nouns extracted with the volcano plot, which are closely related to keywords provided by the experts within the preprints divided into six-month intervals. Terms with the highest score were aggregated. Measuring which compound nouns are used closely over time allows us to identify how the technology evolves without any dependencies on an embedding. The scores of the terms are distributed in the following way.

$$\cdots,\ 0,\ 0.3,\ 0.5,\ 1,\ x,\ 0.5,\ 0.3,\ 0,\ \cdots$$

The $x$ represents the search term. The value around represents the score that each specific term receives. The score is added to an overall counter that matches $x$ and $y$ together. For example, in the following sentence: \textbf{Transformer} is the technology that allows the development of the \textit{large language model} and is the successor of \textit{long short term memory model}. The word in bold represents the search term, and the ones in italics are specialized words. Thus, in this example, \textit{long short term memory model} will increase its score associated to \textbf{transformer} of $0.5$.

This method reduces the delay that appears in the other methods since it uses a closer source, as we use the paper directly. We also have less contamination since we can easily choose the origin of the data. For example, since we use only the cs category of arXiv, we can be almost certain that electrical transformers do not impact our trend on \quotes{Transformer}. The time-frame for our analysis is from January 2017 to September 2021. We start in 2017 as it is the publication year of the Transformer paper~\protect\cite{vaswani_attention_2017}. We output the result in tables where each lines represents the five nouns associated with their score ranked from the highest to the lower score (cf. \autoref{correlation_trends}). A line represent a time-frame of six month.

\begin{figure}
\centering
\includegraphics[width=0.95\textwidth]{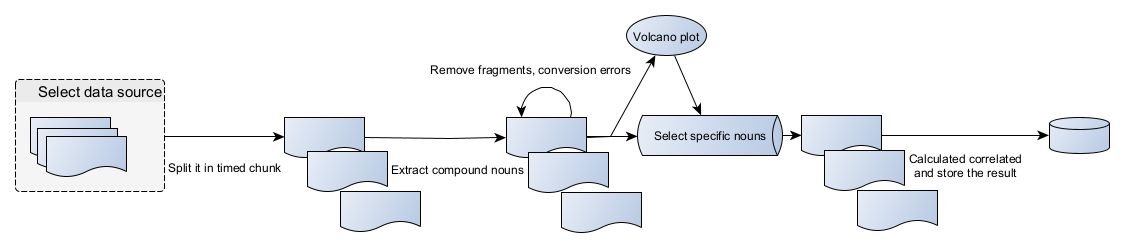}
\caption{Pipeline of the noun \& trends extractor}
\label{fig:nouns_pipeline}
\end{figure}

The overall pipeline is to select the data source (cs categories in arXiv in our case) and split it into chunks of a fixed time gap. Extract the compound nouns. Compare those terms against their appearance frequencies in the bookCorpus (volcano plot). Then, select only the specific one as determined by the volcano plot analysis. Finally, calculate their closeness score in terms of distance and output the one that is the closest to the research technologies (cf. \autoref{fig:nouns_pipeline}).

\section{Results and Discussion}

\subsection{Entity extraction models perform similarly}

Our first result is that in computer science bibliometrics, a variety of entity extraction models perform similarly, with performance being mostly defined by their base architecture and the dataset used to fine-tune them, with the main difference being the number of keywords extracted per document (\autoref{fig:extractor_clusters} and \autoref{table:model_refs}). Given that base, architectures are predominantly BERT and RoBERTa \cite{BERT2018, RoBERTa2019}, and fine-tuning datasets are general texts, such as the ubiquitous Conll03 newswire \cite{Conll2003}. We argue that this means that we cannot expect general LLM-based entity extraction models to perform well on scientific articles. LLM fine-tunes are sensitive to the training data, and among the models tested, only \textit{KBIR-inspec} was fine-tuned using a scientific dataset - INSPEC dataset subsample \cite{INSPEC2003, INSPECBis2019}, consisting of annotated article abstracts from \textit{Computers and Control and Information Technology} journal, published between 1998 to 2002. Given the pace of the evolution of computer science in the last two decades, it is unclear whether fine-tunes based on those abstracts are still relevant for papers today. We believe that the lack of clear clusters in the embedding from \textit{KBIR-inspec} (\autoref{fig:spacy_umap}
\footnote{Results are similar for other embedding and 2D projection algorithms.}
) suggest that they are indeed not relevant anymore\footnote{Well-organized clusters for NERs in spaCy embeddings are a mix of named entities used in theorems in different fields and theorem numbering conventions and are non-informative per se. We provide interactive versions of embeddings in the code repository for readers to investigate them themselves.}. We hence hypothesize that non-LLM-based Yake \cite{yake2020} and spaCy \cite{spaCy2020} keywords and nouns extractors could be essential for addressing these issues, especially given that they already give radically different results compared to LLM-based extractors.

\subsection{Cosine similarity is not well suited to cluster concept-oriented bibliometrics in computer science}

Our second result is that the cosine similarity of embedding of extracted entities does not perform well for concept-oriented bibliometrics in computer science. Even 2D embedding algorithms known for their tendency to overfit to the point of creating local clusters - t-SNE and UMAP \cite{PachterEmbedding2021} - fail to separate different listings (Figures. \ref{fig:spacy_umap}, \ref{fig:glove_umap}), with the exception of NER in common theorem names and naming conventions fragments in spaCy and GPT-2 embeddings. We believe that this supports our claim that current LLM-based entity extraction and comparison pipelines are not applicable for concept-oriented bibliometrics in computer science and, due to their fine-tuning datasets, are likely not applicable in other scientific domains. 
While there is an overlap between different categories - after all, arXiv allows co-listings of hosted papers - we would expect to see such papers as forming an interpolation bridge between domain-specific papers and keywords.

\subsection{Dependence of the embedding space}

Finally, our third result for the entity comparison is that cosine similarity is highly dependent on the algorithm used to embed extracted entities in the vector space (Figures \ref{fig:spacy_umap}, \ref{fig:glove_umap}). While we only present the UMAP embedding here using spaCy and GloVe embeddings, we observe similar results for other embeddings and 2D projection algorithms. Not only do the embeddings drastically modify the representation and similarity of extracted entities, but some embeddings fail to embed numerous entities from some extractors. To illustrate this point, in the code repository of the project, we are providing clustering coefficients (intergroup dispersion vs intra-group dispersion) between different arXiv listings for each entity extractor and each embedding algorithm. We argue that this warrants additional attention when entities extracted from scientific articles are compared using vector space embeddings. While it is not entirely unexpected that embeddings affect the quantitative evaluation of similarity between terms, the impact of that choice is drastic.

\subsection{Performance of noun extraction}

\begin{figure}
    \centering
    \includegraphics[width=0.8\textwidth]{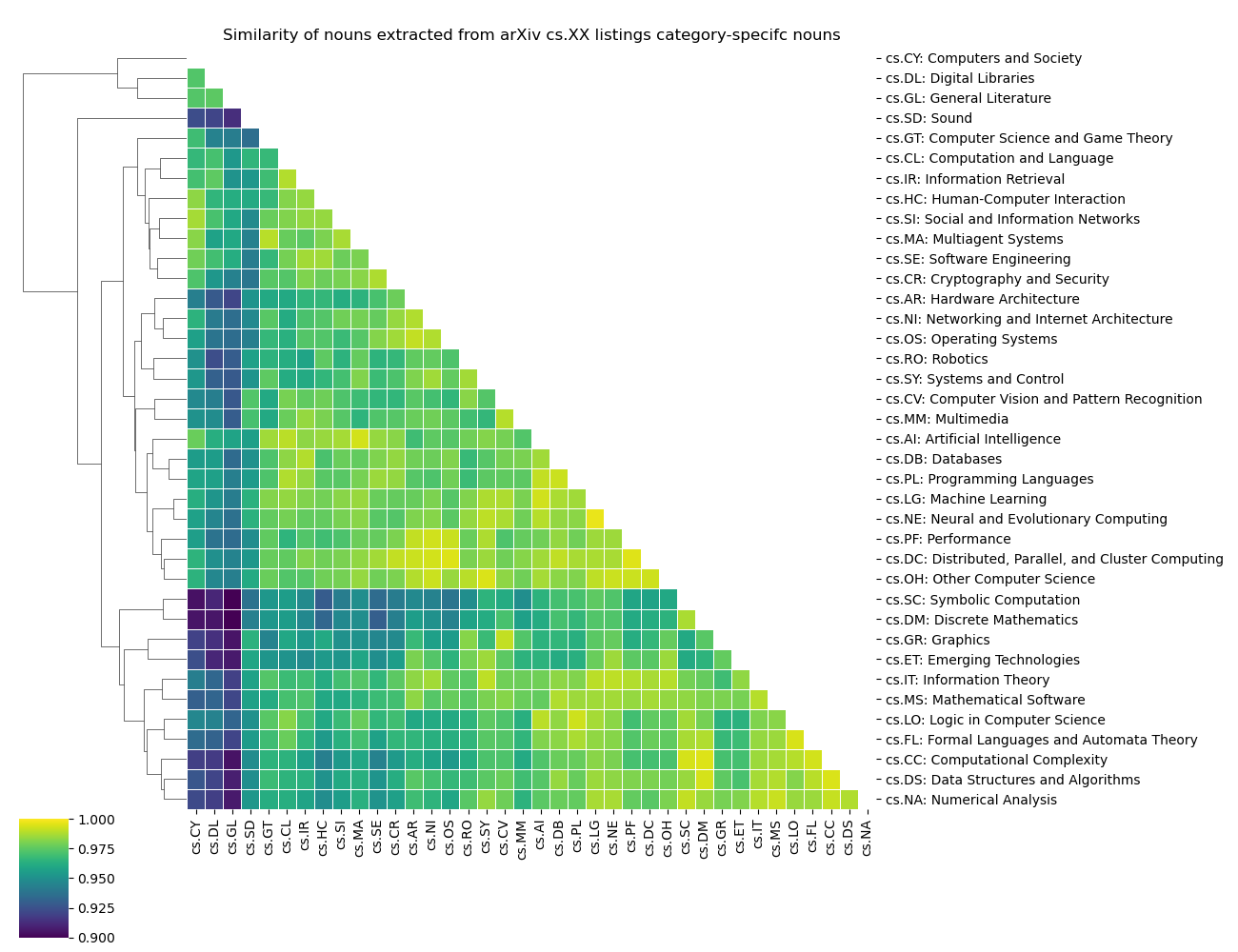}
    \caption{Hierarchical clustering of cs listings based on cosine similarity of the specific nouns extracted in spaCy embedding}
    \label{fig:listing_hierachy}
\end{figure}

\begin{figure}[h]
\centering
\begin{tikzpicture}
    \node[anchor=south west,inner sep=0] (image_with_table) at (0,0) {\includegraphics[width=0.55\textwidth]{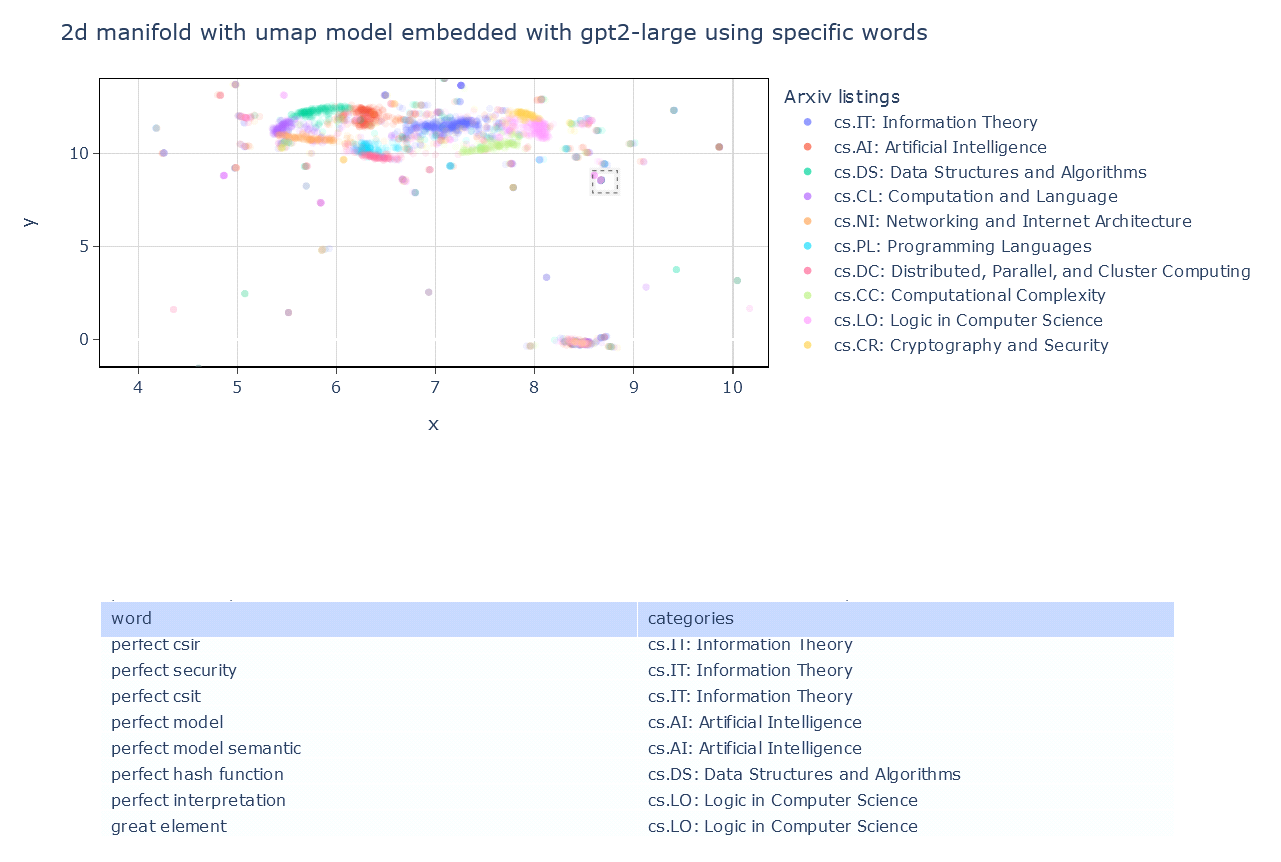}};
    \node[anchor=south east,inner sep=3] (image) at (0,-0.5) {\includegraphics[trim=15 15 300 40,clip,width=0.4\textwidth]{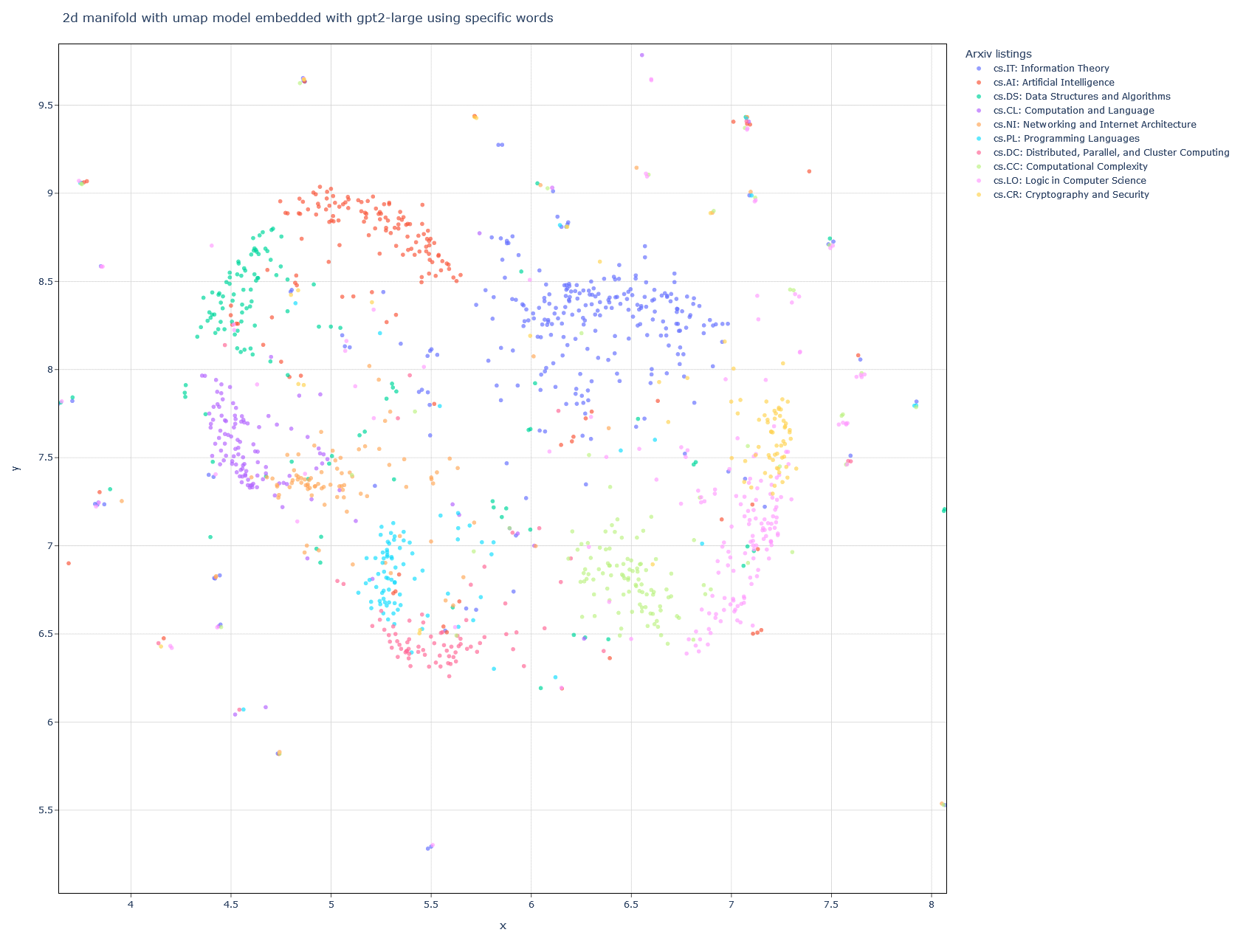}};
    \begin{scope}[x={(image_with_table.south east)},y={(image_with_table.north west)}]
        \draw[black, line width=0.2mm] (0.203, 0.8) rectangle (0.44, 0.895);
        \draw[black, line width=0.2mm] (0.203, 0.895) -- (-0.015, 0.962);
        \draw[black, line width=0.2mm] (0.203, 0.8) -- (-0.015, -0.015);
        \draw[black, line width=0.2mm] (0.462, 0.775) rectangle (0.482, 0.802);
        \draw[black, line width=0.2mm] (0.462, 0.775) -- (0.081, 0.3);
        \draw[black, line width=0.2mm] (0.482, 0.775) -- (0.914, 0.3);
    \end{scope}
\end{tikzpicture}
\caption{2D projection with UMAP of GPT-2 embedding with the specific extracted nouns with a table showing parts of the selected nouns in the figures and a zoom on the main cluster}
\label{fig:gpt2_umap_freq}
\end{figure}

The \autoref{fig:gpt2_umap_freq} illustrates by the presence of cluster that our algorithm is able to extract pertinent compound nouns in the cybersecurity field. Its capabilities to obtain specific knowledge entities are remarkable, especially within a given listing. The specific compound nouns selected with the volcano plot (cf. \autoref{fig:volcano_cr}) correspond well to their corresponding listings. With them, we can observe the similarities among the different listings of the computer science category of arXiv. The \autoref{fig:listing_hierachy} illustrate the differences and similarities among the different listings.

By projecting the specific compound nouns in the GPT-2 embeddings, we can see their ability to cluster in the different listings when applying the UMAP 2d projection (cf. \autoref{fig:gpt2_umap_freq}). The small satellite clusters surrounding the primary one include some highly similar terms present in multiple different listings. The table inside the figure illustrates this. This method offers great opportunities to extract highly specific terms and also recent ones. Some first-class entities can be extracted with this extractor technique. Additional figures can be found in the repository (\url{https://anonymous.4open.science/r/joi-28CE/}).

\subsection{Trends evolution using closeness of words}

While the trends extracted through word usage correlation seem to be informative in their own right, the actual test of the method's usefulness is the ability to anticipate trends that were a posteriori discovered by domain experts. To validate the output of our method, we performed an analysis of trends present in the usage correlation tables retrieved by our method (cf. \autoref{correlation_trends}). 

Overall, the expert evaluation noted that while overall noisy and, in some cases, surprisingly lacking any correlation for fundamental terms, our method could still retrieve early signals of trends that later proved to be critical developments in the LLMs field.

Specifically, for \quotes{Neural Language Model}, our method correctly denoted a transition from conversational research in 2019 to autoencoding-focused models as BERT gained popularity and a transition to tokenization optimization research in  2021. Similarly, for the term \quotes{attention}, we see a spike of application attention mechanism to computer vision models in  2017, as well as the research into the integration of RNNs with the attention mechanism in 2017, as well as the spike in the attention towards the attention triplet in 2021 - tracking ripples from a notable 2020 paper. For the term \quotes{Transformer Model}, we observe the progression from the improvement in embedding algorithms at the end of the second semester for 2018 to improvement in output rating in 2019, to output decoding, too, in 2021, the universality of T5-like models and finally a recent integration between transformers and knowledge graphs through graph neural networks.  

For the \quotes{Large Language Models} term proper, we do see an emergence in late 2019-2020 - around the time it was first coined, and we see a progress towards prompting exploration (meta prompt), to performance on complex tasks (riddlesense), to multilingual benchmarking, tracking some of the major themes in LLMs performance evaluation.

Unfortunately, the method does not only provide useful information. The term \quotes{fine-tuning} provides no association to any terms, and the \quotes{conversational agent} does not provide any informative connections and is overall inconsistent with LLM conversational agents' conceptualization and development timeline.

\section{Conclusion}

The challenges with using the established pipeline in entity extraction and comparison in the context of computer science academic literature we presented above are even more marked for cybersecurity-related topics. In fact, due to the velocity of the field, the new entities rapidly emerge and disappear, meaning that LLM fine-tunes are unlikely to be a reliable solution in that field.

However, the problems we identified with current pipelines for entity extraction and comparison are more general and likely to concern a variety of scientific fields. With our proposed models using compound noun extraction and some statistical analysis to remove common terms, we provide a first step to evolve the entity extractors. Through our results, we see that it performs better than the existing extractor, but it still has some drawbacks that need to be fixed.

We hope that the work presented here will lead to improved practices in the field of knowledge entity extraction and evaluation, as well as motivate further research into better tools. To allow further investigation, we provide in open source all tools used in this project and provide exhaustive and interactive versions of projections in the project repository. We also provide some guidelines about some requirements to improve the models to obtain better tools to monitor the emergent trends.

\subsection{Future work}

When searching for similar words to the search technologies, using embeddings provides poor results due to too recent technologies and too many terms in the compound noun. This is not entirely surprising because embeddings are trained at a fixed time. With the drift of language - especially in technical, rapidly evolving domains- it might have trouble separating terms that came to mean drastically different things, e.g., Deep Machine Learning Vision Models and Deep Machine Learning Language Models. We also need a way to detect and disambiguate common contractions and abbreviations, e.g. (DeepML, RNNs, LSTMs, \ldots). While the use of LLMs might be tempting, here we show that the ones we could use for that purpose locally ($\sim 100-1000\ M$ parameters) actually perform really poorly for conceptual terms extraction in cyber-security.

To obtain a fully deployable tool, there is a need to add the links to papers where the extracted compound nouns come from. This is more software architecture than research work, but it is critical for forecasting validation by an analyst who can manually inspect trends. The aim is to provide a tool that accelerates the detection of emergent technologies. Human feedback is still needed to validate the claims provided by the forecasting algorithm.

There is a need to have better metrics for terms detection than raw distance, e.g., paragraph or even paper co-occurrence.; as well as a better representation of tokens, especially abbreviated ones, given that it made the results non-interpretable here.

\section{Acknowledgments}
AK is supported by the CYD Campus, armasuisse W+T, VBS grant (ARAMIS CYD-C-2020015).

\section{AI Tools Usage}

During the preparation of this work the authors used Grammarly in order to make grammar correction and rephrasing. After using this service, the authors reviewed and edited the content as needed and take full responsibility for the content of the publication.

\bibliography{sample}

\begin{thebibliography}{62}
\providecommand{\natexlab}[1]{#1}
\providecommand{\url}[1]{\texttt{#1}}
\expandafter\ifx\csname urlstyle\endcsname\relax
  \providecommand{\doi}[1]{doi: #1}\else
  \providecommand{\doi}{doi: \begingroup \urlstyle{rm}\Url}\fi

\bibitem[Adner and Levinthal(2002)]{adner2002emergence}
R.~Adner and D.~A. Levinthal.
\newblock The emergence of emerging technologies.
\newblock \emph{California management review}, 45\penalty0 (1):\penalty0 50--66, 2002.

\bibitem[Ai et~al.(2023)Ai, Bai, Cao, Chang, Chen, Chen, Cheng, Dong, Dou, Feng, Gao, Guo, He, Lan, Li, Liu, Lyu, Ma, Ma, Ren, Ren, Wang, Wang, Wen, Wu, Xin, Xu, Yin, Zhang, Zhang, Zhang, Zhang, and Zhu]{ai_information_2023}
Q.~Ai, T.~Bai, Z.~Cao, Y.~Chang, J.~Chen, Z.~Chen, Z.~Cheng, S.~Dong, Z.~Dou, F.~Feng, S.~Gao, J.~Guo, X.~He, Y.~Lan, C.~Li, Y.~Liu, Z.~Lyu, W.~Ma, J.~Ma, Z.~Ren, P.~Ren, Z.~Wang, M.~Wang, J.-R. Wen, L.~Wu, X.~Xin, J.~Xu, D.~Yin, P.~Zhang, F.~Zhang, W.~Zhang, M.~Zhang, and X.~Zhu.
\newblock Information {Retrieval} {Meets} {Large} {Language} {Models}: {A} {Strategic} {Report} from {Chinese} {IR} {Community}, July 2023.
\newblock URL \url{http://arxiv.org/abs/2307.09751}.
\newblock arXiv:2307.09751 [cs].

\bibitem[Anderson(2020)]{anderson_security_2020}
R.~Anderson.
\newblock \emph{Security engineering: a guide to building dependable distributed systems}.
\newblock Wiley, 3 edition, 2020.

\bibitem[Bai et~al.(2022)Bai, Kadavath, Kundu, Askell, Kernion, Jones, Chen, Goldie, Mirhoseini, McKinnon, Chen, Olsson, Olah, Hernandez, Drain, Ganguli, Li, Tran{-}Johnson, Perez, Kerr, Mueller, Ladish, Landau, Ndousse, Lukosiute, Lovitt, Sellitto, Elhage, Schiefer, Mercado, DasSarma, Lasenby, Larson, Ringer, Johnston, Kravec, Showk, Fort, Lanham, Telleen{-}Lawton, Conerly, Henighan, Hume, Bowman, Hatfield{-}Dodds, Mann, Amodei, Joseph, McCandlish, Brown, and Kaplan]{AIBorder2022Anthropic}
Y.~Bai, S.~Kadavath, S.~Kundu, A.~Askell, J.~Kernion, A.~Jones, A.~Chen, A.~Goldie, A.~Mirhoseini, C.~McKinnon, C.~Chen, C.~Olsson, C.~Olah, D.~Hernandez, D.~Drain, D.~Ganguli, D.~Li, E.~Tran{-}Johnson, E.~Perez, J.~Kerr, J.~Mueller, J.~Ladish, J.~Landau, K.~Ndousse, K.~Lukosiute, L.~Lovitt, M.~Sellitto, N.~Elhage, N.~Schiefer, N.~Mercado, N.~DasSarma, R.~Lasenby, R.~Larson, S.~Ringer, S.~Johnston, S.~Kravec, S.~E. Showk, S.~Fort, T.~Lanham, T.~Telleen{-}Lawton, T.~Conerly, T.~Henighan, T.~Hume, S.~R. Bowman, Z.~Hatfield{-}Dodds, B.~Mann, D.~Amodei, N.~Joseph, S.~McCandlish, T.~Brown, and J.~Kaplan.
\newblock Constitutional {AI:} harmlessness from {AI} feedback.
\newblock \emph{CoRR}, abs/2212.08073, 2022.
\newblock \doi{10.48550/arXiv.2212.08073}.
\newblock URL \url{https://doi.org/10.48550/arXiv.2212.08073}.

\bibitem[Bender et~al.(2021)Bender, Gebru, McMillan{-}Major, and Shmitchell]{StochasticParrots2021BenderGebruMitchell}
E.~M. Bender, T.~Gebru, A.~McMillan{-}Major, and S.~Shmitchell.
\newblock On the dangers of stochastic parrots: Can language models be too big?
\newblock In M.~C. Elish, W.~Isaac, and R.~S. Zemel, editors, \emph{FAccT '21: 2021 {ACM} Conference on Fairness, Accountability, and Transparency, Virtual Event / Toronto, Canada, March 3-10, 2021}, pages 610--623. {ACM}, 2021.
\newblock \doi{10.1145/3442188.3445922}.
\newblock URL \url{https://doi.org/10.1145/3442188.3445922}.

\bibitem[Blei et~al.(2003)Blei, Ng, and Jordan]{blei_latent_2003}
D.~M. Blei, A.~Y. Ng, and M.~I. Jordan.
\newblock Latent dirichlet allocation.
\newblock \emph{J. Mach. Learn. Res.}, 3\penalty0 (null):\penalty0 993--1022, Mar. 2003.
\newblock ISSN 1532-4435.

\bibitem[Bojanowski et~al.(2017)Bojanowski, Grave, Joulin, and Mikolov]{FastText2017}
P.~Bojanowski, E.~Grave, A.~Joulin, and T.~Mikolov.
\newblock Enriching word vectors with subword information.
\newblock \emph{Trans. Assoc. Comput. Linguistics}, 5:\penalty0 135--146, 2017.
\newblock \doi{10.1162/tacl\_a\_00051}.
\newblock URL \url{https://doi.org/10.1162/tacl\_a\_00051}.

\bibitem[Campos et~al.(2020)Campos, Mangaravite, Pasquali, Jorge, Nunes, and Jatowt]{yake2020}
R.~Campos, V.~Mangaravite, A.~Pasquali, A.~Jorge, C.~Nunes, and A.~Jatowt.
\newblock Yake! keyword extraction from single documents using multiple local features.
\newblock \emph{Information Sciences}, 509:\penalty0 257--289, 2020.

\bibitem[Chari et~al.(2021)Chari, Banerjee, and Pachter]{PachterEmbedding2021}
T.~Chari, J.~Banerjee, and L.~Pachter.
\newblock The specious art of single-cell genomics.
\newblock \emph{BioRxiv}, pages 2021--08, 2021.

\bibitem[Chen et~al.(2021)Chen, Xie, Li, and Cheng]{chen2021topic}
X.~Chen, H.~Xie, Z.~Li, and G.~Cheng.
\newblock Topic analysis and development in knowledge graph research: A bibliometric review on three decades.
\newblock \emph{Neurocomputing}, 461:\penalty0 497--515, 2021.

\bibitem[Chumnumpan and Shi(2019)]{chumnumpan2019understanding}
P.~Chumnumpan and X.~Shi.
\newblock Understanding new products’ market performance using google trends.
\newblock \emph{Australasian marketing journal}, 27\penalty0 (2):\penalty0 91--103, 2019.

\bibitem[Clark et~al.(2020)Clark, Luong, Le, and Manning]{Electra2020}
K.~Clark, M.~Luong, Q.~V. Le, and C.~D. Manning.
\newblock {ELECTRA:} pre-training text encoders as discriminators rather than generators.
\newblock In \emph{8th International Conference on Learning Representations, {ICLR} 2020, Addis Ababa, Ethiopia, April 26-30, 2020}. OpenReview.net, 2020.
\newblock URL \url{https://openreview.net/forum?id=r1xMH1BtvB}.

\bibitem[Daim and Yalçin(2022)]{daim_digital_2022}
T.~Daim and H.~Yalçin.
\newblock \emph{Digital transformations: new tools and methods for mining technological intelligence}.
\newblock Edward Elgar Publishing, 2022.
\newblock ISBN 978-1-78990-862-6.

\bibitem[Daim et~al.(2016)Daim, Chiavetta, Porter, and Saritas]{daim_anticipating_2016}
T.~U. Daim, D.~Chiavetta, A.~L. Porter, and O.~Saritas.
\newblock \emph{Anticipating future innovation pathways through large data analysis}.
\newblock Springer, 2016.
\newblock ISBN 978-3-319-39056-7.

\bibitem[Devlin et~al.(2018)Devlin, Chang, Lee, and Toutanova]{BERT2018}
J.~Devlin, M.~Chang, K.~Lee, and K.~Toutanova.
\newblock {BERT:} pre-training of deep bidirectional transformers for language understanding.
\newblock \emph{CoRR}, abs/1810.04805, 2018.
\newblock URL \url{http://arxiv.org/abs/1810.04805}.

\bibitem[Donthu et~al.(2021)Donthu, Kumar, Mukherjee, Pandey, and Lim]{donthu_how_2021}
N.~Donthu, S.~Kumar, D.~Mukherjee, N.~Pandey, and W.~M. Lim.
\newblock How to conduct a bibliometric analysis: {An} overview and guidelines.
\newblock \emph{Journal of Business Research}, 133:\penalty0 285--296, Sept. 2021.
\newblock ISSN 0148-2963.
\newblock \doi{10.1016/j.jbusres.2021.04.070}.
\newblock URL \url{https://www.sciencedirect.com/science/article/pii/S0148296321003155}.

\bibitem[Dunn et~al.(2022)Dunn, Dagdelen, Walker, Lee, Rosen, Ceder, Persson, and Jain]{dunn2022structured}
A.~Dunn, J.~Dagdelen, N.~Walker, S.~Lee, A.~S. Rosen, G.~Ceder, K.~Persson, and A.~Jain.
\newblock Structured information extraction from complex scientific text with fine-tuned large language models.
\newblock \emph{arXiv preprint arXiv:2212.05238}, 2022.

\bibitem[Ganguli et~al.(2022)Ganguli, Hernandez, Lovitt, Askell, Bai, Chen, Conerly, DasSarma, Drain, Elhage, Showk, Fort, Hatfield{-}Dodds, Henighan, Johnston, Jones, Joseph, Kernian, Kravec, Mann, Nanda, Ndousse, Olsson, Amodei, Brown, Kaplan, McCandlish, Olah, Amodei, and Clark]{UnpredictabilityOfLLMs2022Anthropic}
D.~Ganguli, D.~Hernandez, L.~Lovitt, A.~Askell, Y.~Bai, A.~Chen, T.~Conerly, N.~DasSarma, D.~Drain, N.~Elhage, S.~E. Showk, S.~Fort, Z.~Hatfield{-}Dodds, T.~Henighan, S.~Johnston, A.~Jones, N.~Joseph, J.~Kernian, S.~Kravec, B.~Mann, N.~Nanda, K.~Ndousse, C.~Olsson, D.~Amodei, T.~Brown, J.~Kaplan, S.~McCandlish, C.~Olah, D.~Amodei, and J.~Clark.
\newblock Predictability and surprise in large generative models.
\newblock In \emph{FAccT '22: 2022 {ACM} Conference on Fairness, Accountability, and Transparency, Seoul, Republic of Korea, June 21 - 24, 2022}, pages 1747--1764. {ACM}, 2022.
\newblock \doi{10.1145/3531146.3533229}.
\newblock URL \url{https://doi.org/10.1145/3531146.3533229}.

\bibitem[Gao et~al.(2021)Gao, Zhang, and Liu]{gao_data_2021}
C.~Gao, X.~Zhang, and H.~Liu.
\newblock Data and knowledge-driven named entity recognition for cyber security.
\newblock \emph{Cybersecurity}, 4\penalty0 (1):\penalty0 9, May 2021.
\newblock ISSN 2523-3246.
\newblock \doi{10.1186/s42400-021-00072-y}.
\newblock URL \url{https://doi.org/10.1186/s42400-021-00072-y}.

\bibitem[Goyal et~al.(2021)Goyal, Du, Ott, Anantharaman, and Conneau]{XLMRoBERTaXL2021}
N.~Goyal, J.~Du, M.~Ott, G.~Anantharaman, and A.~Conneau.
\newblock Larger-scale transformers for multilingual masked language modeling.
\newblock In A.~Rogers, I.~Calixto, I.~Vulic, N.~Saphra, N.~Kassner, O.~Camburu, T.~Bansal, and V.~Shwartz, editors, \emph{Proceedings of the 6th Workshop on Representation Learning for NLP, RepL4NLP@ACL-IJCNLP 2021, Online, August 6, 2021}, pages 29--33. Association for Computational Linguistics, 2021.
\newblock \doi{10.18653/v1/2021.repl4nlp-1.4}.
\newblock URL \url{https://doi.org/10.18653/v1/2021.repl4nlp-1.4}.

\bibitem[Grootendorst(2020)]{grootendorst2020keybert}
M.~Grootendorst.
\newblock Keybert: Minimal keyword extraction with bert., 2020.
\newblock URL \url{https://doi.org/10.5281/zenodo.4461265}.

\bibitem[Hoffmann et~al.(2022)Hoffmann, Borgeaud, Mensch, Buchatskaya, Cai, Rutherford, de~Las~Casas, Hendricks, Welbl, Clark, Hennigan, Noland, Millican, van~den Driessche, Damoc, Guy, Osindero, Simonyan, Elsen, Rae, Vinyals, and Sifre]{ComputeOptimalLLMs2022Google}
J.~Hoffmann, S.~Borgeaud, A.~Mensch, E.~Buchatskaya, T.~Cai, E.~Rutherford, D.~de~Las~Casas, L.~A. Hendricks, J.~Welbl, A.~Clark, T.~Hennigan, E.~Noland, K.~Millican, G.~van~den Driessche, B.~Damoc, A.~Guy, S.~Osindero, K.~Simonyan, E.~Elsen, J.~W. Rae, O.~Vinyals, and L.~Sifre.
\newblock Training compute-optimal large language models.
\newblock \emph{CoRR}, abs/2203.15556, 2022.
\newblock \doi{10.48550/arXiv.2203.15556}.
\newblock URL \url{https://doi.org/10.48550/arXiv.2203.15556}.

\bibitem[Honnibal et~al.(2020)Honnibal, Montani, Van~Landeghem, and Boyd]{spaCy2020}
M.~Honnibal, I.~Montani, S.~Van~Landeghem, and A.~Boyd.
\newblock {spaCy: Industrial-strength Natural Language Processing in Python}.
\newblock \emph{Zenodo}, 2020.
\newblock \doi{10.5281/zenodo.1212303}.

\bibitem[Hovy et~al.(2006)Hovy, Marcus, Palmer, Ramshaw, and Weischedel]{Ontonotes2006}
E.~H. Hovy, M.~P. Marcus, M.~Palmer, L.~A. Ramshaw, and R.~M. Weischedel.
\newblock Ontonotes: The 90{\%} solution.
\newblock In R.~C. Moore, J.~A. Bilmes, J.~Chu{-}Carroll, and M.~Sanderson, editors, \emph{Human Language Technology Conference of the North American Chapter of the Association of Computational Linguistics, Proceedings, June 4-9, 2006, New York, New York, {USA}}. The Association for Computational Linguistics, 2006.
\newblock URL \url{https://aclanthology.org/N06-2015/}.

\bibitem[Hulth(2003)]{INSPEC2003}
A.~Hulth.
\newblock Improved automatic keyword extraction given more linguistic knowledge.
\newblock In \emph{Proceedings of the 2003 conference on Empirical methods in natural language processing}, pages 216--223, 2003.

\bibitem[Ishizaki and Kaufer(2012)]{docusco2012}
S.~Ishizaki and D.~Kaufer.
\newblock Computer-aided rhetorical analysis.
\newblock In \emph{Applied natural language processing: Identification, investigation and resolution}, pages 276--296. IGI Global, 2012.
\newblock URL \url{https://www.igi-global.com/chapter/content/61054}.

\bibitem[Jun et~al.(2018{\natexlab{a}})Jun, Yoo, and Choi]{jun2018ten}
S.-P. Jun, H.~S. Yoo, and S.~Choi.
\newblock Ten years of research change using google trends: From the perspective of big data utilizations and applications.
\newblock \emph{Technological forecasting and social change}, 130:\penalty0 69--87, 2018{\natexlab{a}}.

\bibitem[Jun et~al.(2018{\natexlab{b}})Jun, Yoo, and Choi]{jun_ten_2018}
S.-P. Jun, H.~S. Yoo, and S.~Choi.
\newblock Ten years of research change using {Google} {Trends}: {From} the perspective of big data utilizations and applications.
\newblock \emph{Technological Forecasting and Social Change}, 130:\penalty0 69--87, May 2018{\natexlab{b}}.
\newblock ISSN 0040-1625.
\newblock \doi{10.1016/j.techfore.2017.11.009}.
\newblock URL \url{https://www.sciencedirect.com/science/article/pii/S0040162517315536}.

\bibitem[Kaplan et~al.(2020)Kaplan, McCandlish, Henighan, Brown, Chess, Child, Gray, Radford, Wu, and Amodei]{ScalingLawsLLM2020OpenAI}
J.~Kaplan, S.~McCandlish, T.~Henighan, T.~B. Brown, B.~Chess, R.~Child, S.~Gray, A.~Radford, J.~Wu, and D.~Amodei.
\newblock Scaling laws for neural language models.
\newblock \emph{CoRR}, abs/2001.08361, 2020.
\newblock URL \url{https://arxiv.org/abs/2001.08361}.

\bibitem[Klepper(1997)]{klepper1997industry}
S.~Klepper.
\newblock Industry life cycles.
\newblock \emph{Industrial and corporate change}, 6\penalty0 (1):\penalty0 145--182, 1997.

\bibitem[K{\"{o}}pf et~al.(2023)K{\"{o}}pf, Kilcher, von R{\"{u}}tte, Anagnostidis, Tam, Stevens, Barhoum, Duc, Stanley, Nagyfi, ES, Suri, Glushkov, Dantuluri, Maguire, Schuhmann, Nguyen, and Mattick]{LIAON2023OpenAssistantConvos}
A.~K{\"{o}}pf, Y.~Kilcher, D.~von R{\"{u}}tte, S.~Anagnostidis, Z.~Tam, K.~Stevens, A.~Barhoum, N.~M. Duc, O.~Stanley, R.~Nagyfi, S.~ES, S.~Suri, D.~Glushkov, A.~Dantuluri, A.~Maguire, C.~Schuhmann, H.~Nguyen, and A.~Mattick.
\newblock Openassistant conversations - democratizing large language model alignment.
\newblock \emph{CoRR}, abs/2304.07327, 2023.
\newblock \doi{10.48550/arXiv.2304.07327}.
\newblock URL \url{https://doi.org/10.48550/arXiv.2304.07327}.

\bibitem[Kucharavy et~al.(2023)Kucharavy, Schillaci, Mar{\'e}chal, W{\"u}rsch, Dolamic, Sabonnadiere, David, Mermoud, and Lenders]{kucharavy_fundamentals_2023}
A.~Kucharavy, Z.~Schillaci, L.~Mar{\'e}chal, M.~W{\"u}rsch, L.~Dolamic, R.~Sabonnadiere, D.~P. David, A.~Mermoud, and V.~Lenders.
\newblock Fundamentals of {Generative} {Large} {Language} {Models} and {Perspectives} in {Cyber}-{Defense}, Mar. 2023.
\newblock URL \url{http://arxiv.org/abs/2303.12132}.
\newblock arXiv:2303.12132 [cs].

\bibitem[Kulkarni et~al.(2022)Kulkarni, Mahata, Arora, and Bhowmik]{kbir22}
M.~Kulkarni, D.~Mahata, R.~Arora, and R.~Bhowmik.
\newblock Learning rich representation of keyphrases from text.
\newblock In M.~Carpuat, M.~de~Marneffe, and I.~V.~M. Ru{\'{\i}}z, editors, \emph{Findings of the Association for Computational Linguistics: {NAACL} 2022, Seattle, WA, United States, July 10-15, 2022}, pages 891--906. Association for Computational Linguistics, 2022.
\newblock \doi{10.18653/v1/2022.findings-naacl.67}.
\newblock URL \url{https://doi.org/10.18653/v1/2022.findings-naacl.67}.

\bibitem[Li et~al.(2020)Li, Sun, Han, and Li]{li2020survey}
J.~Li, A.~Sun, J.~Han, and C.~Li.
\newblock A survey on deep learning for named entity recognition.
\newblock \emph{IEEE Transactions on Knowledge and Data Engineering}, 34\penalty0 (1):\penalty0 50--70, 2020.

\bibitem[Liu et~al.(2019)Liu, Ott, Goyal, Du, Joshi, Chen, Levy, Lewis, Zettlemoyer, and Stoyanov]{RoBERTa2019}
Y.~Liu, M.~Ott, N.~Goyal, J.~Du, M.~Joshi, D.~Chen, O.~Levy, M.~Lewis, L.~Zettlemoyer, and V.~Stoyanov.
\newblock Roberta: {A} robustly optimized {BERT} pretraining approach.
\newblock \emph{CoRR}, abs/1907.11692, 2019.
\newblock URL \url{http://arxiv.org/abs/1907.11692}.

\bibitem[Marujo et~al.(2012)Marujo, Gershman, Carbonell, Frederking, and Neto]{KPCrowd2012}
L.~Marujo, A.~Gershman, J.~G. Carbonell, R.~E. Frederking, and J.~P. Neto.
\newblock Supervised topical key phrase extraction of news stories using crowdsourcing, light filtering and co-reference normalization.
\newblock In N.~Calzolari, K.~Choukri, T.~Declerck, M.~U. Dogan, B.~Maegaard, J.~Mariani, J.~Odijk, and S.~Piperidis, editors, \emph{Proceedings of the Eighth International Conference on Language Resources and Evaluation, {LREC} 2012, Istanbul, Turkey, May 23-25, 2012}, pages 399--403. European Language Resources Association {(ELRA)}, 2012.
\newblock URL \url{http://www.lrec-conf.org/proceedings/lrec2012/summaries/672.html}.

\bibitem[McInnes and Healy(2018)]{UMAP2018}
L.~McInnes and J.~Healy.
\newblock {UMAP:} uniform manifold approximation and projection for dimension reduction.
\newblock \emph{CoRR}, abs/1802.03426, 2018.
\newblock URL \url{http://arxiv.org/abs/1802.03426}.

\bibitem[Mikolov et~al.(2013)Mikolov, Chen, Corrado, and Dean]{Word2Vec2013}
T.~Mikolov, K.~Chen, G.~Corrado, and J.~Dean.
\newblock Efficient estimation of word representations in vector space.
\newblock In Y.~Bengio and Y.~LeCun, editors, \emph{1st International Conference on Learning Representations, {ICLR} 2013, Scottsdale, Arizona, USA, May 2-4, 2013, Workshop Track Proceedings}, 2013.
\newblock URL \url{http://arxiv.org/abs/1301.3781}.

\bibitem[OpenAI(2023)]{GPT42023OpenAI}
OpenAI.
\newblock Gpt-4 technical report.
\newblock \emph{CoRR}, abs/2303.08774, 2023.
\newblock URL \url{https://arxiv.org/abs/2303.08774}.

\bibitem[Ouyang et~al.(2022)Ouyang, Wu, Jiang, Almeida, Wainwright, Mishkin, Zhang, Agarwal, Slama, Ray, Schulman, Hilton, Kelton, Miller, Simens, Askell, Welinder, Christiano, Leike, and Lowe]{InstructGPT2022OpenAI}
L.~Ouyang, J.~Wu, X.~Jiang, D.~Almeida, C.~L. Wainwright, P.~Mishkin, C.~Zhang, S.~Agarwal, K.~Slama, A.~Ray, J.~Schulman, J.~Hilton, F.~Kelton, L.~Miller, M.~Simens, A.~Askell, P.~Welinder, P.~F. Christiano, J.~Leike, and R.~Lowe.
\newblock Training language models to follow instructions with human feedback.
\newblock \emph{CoRR}, abs/2203.02155, 2022.
\newblock \doi{10.48550/arXiv.2203.02155}.
\newblock URL \url{https://doi.org/10.48550/arXiv.2203.02155}.

\bibitem[Pennington et~al.(2014)Pennington, Socher, and Manning]{GloVe2014}
J.~Pennington, R.~Socher, and C.~D. Manning.
\newblock Glove: Global vectors for word representation.
\newblock In A.~Moschitti, B.~Pang, and W.~Daelemans, editors, \emph{Proceedings of the 2014 Conference on Empirical Methods in Natural Language Processing, {EMNLP} 2014, October 25-29, 2014, Doha, Qatar, {A} meeting of SIGDAT, a Special Interest Group of the {ACL}}, pages 1532--1543. {ACL}, 2014.
\newblock \doi{10.3115/v1/d14-1162}.
\newblock URL \url{https://doi.org/10.3115/v1/d14-1162}.

\bibitem[Percia~David(2020)]{percia2020three}
D.~Percia~David.
\newblock \emph{Three Articles on the Economics of Information-Systems Defense Capability. Material-, Human-, and Knowledge-Resources Acquisition for Critical Infrastructures}.
\newblock PhD thesis, Universit{\'e} de Lausanne, HEC Lausanne, 2020.

\bibitem[Percia~David et~al.(2023)Percia~David, Mar{\'e}chal, Lacube, Gillard, Tsesmelis, Maillart, and Mermoud]{david2023measuring}
D.~Percia~David, L.~Mar{\'e}chal, W.~Lacube, S.~Gillard, M.~Tsesmelis, T.~Maillart, and A.~Mermoud.
\newblock Measuring security development in information technologies: A scientometric framework using arxiv e-prints.
\newblock \emph{Technological Forecasting and Social Change}, 188:\penalty0 122316, 2023.

\bibitem[Perez(2010)]{perez2010technological}
C.~Perez.
\newblock Technological revolutions and techno-economic paradigms.
\newblock \emph{Cambridge journal of economics}, 34\penalty0 (1):\penalty0 185--202, 2010.

\bibitem[Peters et~al.(2018)Peters, Neumann, Iyyer, Gardner, Clark, Lee, and Zettlemoyer]{ElMo2018}
M.~E. Peters, M.~Neumann, M.~Iyyer, M.~Gardner, C.~Clark, K.~Lee, and L.~Zettlemoyer.
\newblock Deep contextualized word representations.
\newblock In M.~A. Walker, H.~Ji, and A.~Stent, editors, \emph{Proceedings of the 2018 Conference of the North American Chapter of the Association for Computational Linguistics: Human Language Technologies, {NAACL-HLT} 2018, New Orleans, Louisiana, USA, June 1-6, 2018, Volume 1 (Long Papers)}, pages 2227--2237. Association for Computational Linguistics, 2018.
\newblock \doi{10.18653/v1/n18-1202}.
\newblock URL \url{https://doi.org/10.18653/v1/n18-1202}.

\bibitem[Petroni et~al.(2019)Petroni, Rockt{\"a}schel, Lewis, Bakhtin, Wu, Miller, and Riedel]{petroni2019language}
F.~Petroni, T.~Rockt{\"a}schel, P.~Lewis, A.~Bakhtin, Y.~Wu, A.~H. Miller, and S.~Riedel.
\newblock Language models as knowledge bases?
\newblock \emph{arXiv preprint arXiv:1909.01066}, 2019.

\bibitem[Priem et~al.(2022)Priem, Piwowar, and Orr]{priem2022openalex}
J.~Priem, H.~Piwowar, and R.~Orr.
\newblock Openalex: A fully-open index of scholarly works, authors, venues, institutions, and concepts.
\newblock \emph{arXiv preprint arXiv:2205.01833}, 2022.

\bibitem[Radford et~al.(2019)Radford, Wu, Child, Luan, Amodei, Sutskever, et~al.]{GPT22019RadfordSutskeverOpenAI}
A.~Radford, J.~Wu, R.~Child, D.~Luan, D.~Amodei, I.~Sutskever, et~al.
\newblock Language models are unsupervised multitask learners.
\newblock \emph{OpenAI blog}, 1\penalty0 (8):\penalty0 9, 2019.

\bibitem[Rogers(2010)]{rogers_diffusion_2010}
E.~M. Rogers.
\newblock \emph{Diffusion of innovations}.
\newblock Simon and Schuster, 2010.

\bibitem[Safder and Hassan(2019)]{safder2019bibliometric}
I.~Safder and S.-U. Hassan.
\newblock Bibliometric-enhanced information retrieval: a novel deep feature engineering approach for algorithm searching from full-text publications.
\newblock \emph{Scientometrics}, 119:\penalty0 257--277, 2019.

\bibitem[Sahrawat et~al.(2019)Sahrawat, Mahata, Kulkarni, Zhang, Gosangi, Stent, Sharma, Kumar, Shah, and Zimmermann]{INSPECBis2019}
D.~Sahrawat, D.~Mahata, M.~Kulkarni, H.~Zhang, R.~Gosangi, A.~Stent, A.~Sharma, Y.~Kumar, R.~R. Shah, and R.~Zimmermann.
\newblock Keyphrase extraction from scholarly articles as sequence labeling using contextualized embeddings.
\newblock \emph{CoRR}, abs/1910.08840, 2019.
\newblock URL \url{http://arxiv.org/abs/1910.08840}.

\bibitem[Sang and Meulder(2003)]{Conll2003}
E.~F. T.~K. Sang and F.~D. Meulder.
\newblock Introduction to the conll-2003 shared task: Language-independent named entity recognition.
\newblock In W.~Daelemans and M.~Osborne, editors, \emph{Proceedings of the Seventh Conference on Natural Language Learning, CoNLL 2003, Held in cooperation with {HLT-NAACL} 2003, Edmonton, Canada, May 31 - June 1, 2003}, pages 142--147. {ACL}, 2003.
\newblock URL \url{https://aclanthology.org/W03-0419/}.

\bibitem[Sanh et~al.(2019)Sanh, Debut, Chaumond, and Wolf]{DistilBERT2019}
V.~Sanh, L.~Debut, J.~Chaumond, and T.~Wolf.
\newblock Distilbert, a distilled version of {BERT:} smaller, faster, cheaper and lighter.
\newblock \emph{CoRR}, abs/1910.01108, 2019.
\newblock URL \url{http://arxiv.org/abs/1910.01108}.

\bibitem[Srivastava et~al.(2023)Srivastava, Paul, and Gupta]{srivastava_study_2023}
S.~Srivastava, B.~Paul, and D.~Gupta.
\newblock Study of {Word} {Embeddings} for {Enhanced} {Cyber} {Security} {Named} {Entity} {Recognition}.
\newblock \emph{Procedia Computer Science}, 218:\penalty0 449--460, Jan. 2023.
\newblock ISSN 1877-0509.
\newblock \doi{10.1016/j.procs.2023.01.027}.
\newblock URL \url{https://www.sciencedirect.com/science/article/pii/S1877050923000273}.

\bibitem[Sun et~al.(2020)Sun, Hu, Li, Li, Li, and Chi]{sun_review_2020}
C.~Sun, L.~Hu, S.~Li, T.~Li, H.~Li, and L.~Chi.
\newblock A {Review} of {Unsupervised} {Keyphrase} {Extraction} {Methods} {Using} {Within}-{Collection} {Resources}.
\newblock \emph{Symmetry}, 12\penalty0 (11):\penalty0 1864, Nov. 2020.
\newblock ISSN 2073-8994.
\newblock \doi{10.3390/sym12111864}.
\newblock URL \url{https://www.mdpi.com/2073-8994/12/11/1864}.
\newblock Number: 11 Publisher: Multidisciplinary Digital Publishing Institute.

\bibitem[Ushio and Camacho{-}Collados(2021)]{XLMRoBERTaBis2021}
A.~Ushio and J.~Camacho{-}Collados.
\newblock {T-NER:} an all-round python library for transformer-based named entity recognition.
\newblock In D.~Gkatzia and D.~Seddah, editors, \emph{Proceedings of the 16th Conference of the European Chapter of the Association for Computational Linguistics: System Demonstrations, {EACL} 2021, Online, April 19-23, 2021}, pages 53--62. Association for Computational Linguistics, 2021.
\newblock \doi{10.18653/v1/2021.eacl-demos.7}.
\newblock URL \url{https://doi.org/10.18653/v1/2021.eacl-demos.7}.

\bibitem[Van~der Maaten and Hinton(2008)]{TSNE2008}
L.~Van~der Maaten and G.~Hinton.
\newblock Visualizing data using t-sne.
\newblock \emph{Journal of machine learning research}, 9\penalty0 (11), 2008.

\bibitem[Vaswani et~al.(2017)Vaswani, Shazeer, Parmar, Uszkoreit, Jones, Gomez, Kaiser, and Polosukhin]{vaswani_attention_2017}
A.~Vaswani, N.~Shazeer, N.~Parmar, J.~Uszkoreit, L.~Jones, A.~N. Gomez, L.~Kaiser, and I.~Polosukhin.
\newblock Attention {Is} {All} {You} {Need}, Dec. 2017.
\newblock URL \url{http://arxiv.org/abs/1706.03762}.
\newblock arXiv:1706.03762 [cs].

\bibitem[West(2020)]{gtab}
R.~West.
\newblock Calibration of google trends time series.
\newblock \emph{CoRR}, abs/2007.13861, 2020.
\newblock URL \url{https://arxiv.org/abs/2007.13861}.

\bibitem[Woloszko(2020)]{woloszko2020tracking}
N.~Woloszko.
\newblock Tracking activity in real time with google trends.
\newblock \emph{OECD Economics Department Working Papers}, 2020.

\bibitem[Zhang et~al.(2018)Zhang, Lu, Liu, Liu, Porter, Chen, and Zhang]{zhang2018does}
Y.~Zhang, J.~Lu, F.~Liu, Q.~Liu, A.~Porter, H.~Chen, and G.~Zhang.
\newblock Does deep learning help topic extraction? a kernel k-means clustering method with word embedding.
\newblock \emph{Journal of Informetrics}, 12\penalty0 (4):\penalty0 1099--1117, 2018.

\bibitem[{‘OurResearch, Org.’}(2023)]{OpenAlex2023}
{‘OurResearch, Org.’}.
\newblock Historical citation data, 2023.
\newblock URL \url{https://docs.openalex.org/api-entities/works}.
\newblock Accessed on 16th of March 2023.

\end{thebibliography}

\clearpage
\appendix
\appendixpage
\section{Evolution of most correlated term over time for several target technologies} \label{correlation_trends}

\begin{table}[h]
\centering
\scriptsize
\begin{tabular}{ ||m{1cm}|m{2.2cm}|m{2.2cm}|m{2.2cm}|m{2.2cm}|m{2.2cm}|| }
\hline
\multicolumn{6}{||c||}{Trends of \quotes{neural language model} term in arXiv} \\
\hline
Semester & Noun 1 & Noun 2 & Noun 3 & Noun 4 & Noun 5 \\
\hline
\hline
2017 S1 & nmt same & augment neural language model & previous output representation & value separation & backtranslate monolingual corpora \\
 & (8.5) & (5.2) & (4.5) & (3.0) & (2.8) \\
\hline
2017 S2 & cross-lingual phonetic representation learning & domain semantic parser & typology prediction & uriel language typology database & good benchmark task \\
 & (1.0) & (1.0) & (1.0) & (1.0) & (1.0) \\
\hline
2018 S1 & neural network translation model & reuse word & cross-lingual phonetic representation learning & david mortensen & perform joint inference \\
 & (1.0) & (1.0) & (1.0) & (1.0) & (1.0) \\
\hline
2018 S2 & segmental language model & previous segmentation decision & underlie segmentation & transformative & share lstm network \\
 & (1.5) & (1.5) & (1.5) & (1.0) & (1.0) \\
\hline
2019 S1 & dtfit & example candidate & synthetic note & dobj centroid & same training corpora \\
 & (1.0) & (1.0) & (1.0) & (1.0) & (1.0) \\
\hline
2019 S2 & conversational style matching & structured discriminator & translation lattice & modern nmt architecture & learn pair \\
 & (1.5) & (1.1) & (1.0) & (1.0) & (1.0) \\
\hline
2020 S1 & generate phrase representation & recent pre-training method & raw text corpora & continual pre-training & non-textual input \\
 & (1.5) & (1.0) & (1.0) & (1.0) & (1.0) \\
\hline
2020 S2 & weird yankovic & 21st nordic conference & parallel effort & xinyu dai & base lstm language model \\
 & (1.5) & (1.0) & (1.0) & (1.0) & (1.0) \\
\hline
2021 S1 & cross-lingual phonetic representation learning & david mortensen & class member variable & aware seq2seq model & subinstruction \\
 & (1.0) & (1.0) & (1.0) & (1.0) & (1.0) \\
\hline
2021 S2 & temporary ambiguity & good tokenisation & possible tokenisation & multimodal semantic representation & syntactic frame extension model \\
 & (2.6) & (1.3) & (1.3) & (1.0) & (1.0) \\
\hline
\end{tabular}
\caption{Evolution the target term \quotes{neural language model}. Based on terms with a score higher than one, we can see a transition from conversational research in 2019 to auto-encoding model in the first semester of 2020 and finally to tokenization question in the second semester of 2021. In overall difficult to extract due to a lot of noise due to the single-occurence connections.
 }
\end{table}

\begin{table}[h]
\centering
\scriptsize
\begin{tabular}{ ||m{1cm}|m{2.2cm}|m{2.2cm}|m{2.2cm}|m{2.2cm}|m{2.2cm}|| }
\hline
\multicolumn{6}{||c||}{Trends of \quotes{deep neural language model} term in arXiv} \\
\hline
Semester & Noun 1 & Noun 2 & Noun 3 & Noun 4 & Noun 5 \\
\hline
\hline
2017 S1 &  &  &  &  &  \\
 & (nan) & (nan) & (nan) & (nan) & (nan) \\
\hline
2017 S2 &  &  &  &  &  \\
 & (nan) & (nan) & (nan) & (nan) & (nan) \\
\hline
2018 S1 & k-hop & orimaye & regularize nonnegative matrix factorization & first derivative saliency & accurate neural model \\
 & (1.0) & (1.0) & (1.0) & (1.0) & (1.0) \\
\hline
2018 S2 & miss relation type prediction & dolores embedding & dolore embedding & contextdependent representation & deep word representation \\
 & (1.5) & (1.5) & (1.3) & (1.0) & (1.0) \\
\hline
2019 S1 &  &  &  &  &  \\
 & (nan) & (nan) & (nan) & (nan) & (nan) \\
\hline
2019 S2 & outperform bo w & little information content & single semantic concept & deep contextualized representation & small contextual window \\
 & (1.0) & (1.0) & (1.0) & (1.0) & (1.0) \\
\hline
2020 S1 & simpletransformer & nltk tool & translate tweet & analysis quality & sen wave \\
 & (1.0) & (1.0) & (0.5) & (0.5) & (0.3) \\
\hline
2020 S2 & harnack & order other & conditional program generation & source domain error & bergh \\
 & (1.0) & (1.0) & (1.0) & (1.0) & (0.5) \\
\hline
2021 S1 & non-degenerate gaussian measure & judgments & permutation language & predict covariance matrix & munzner \\
 & (3.0) & (1.0) & (1.0) & (1.0) & (1.0) \\
\hline
2021 S2 &  &  &  &  &  \\
 & (nan) & (nan) & (nan) & (nan) & (nan) \\
\hline
\end{tabular}
\caption{Evolution of the target term \quotes{deep neural language model}. The resulting table do not provide informative material to analyse the evolution. Most likely to a term to specific.
}
\end{table}

\begin{table}[h]
\centering
\scriptsize
\begin{tabular}{ ||m{1cm}|m{2.2cm}|m{2.2cm}|m{2.2cm}|m{2.2cm}|m{2.2cm}|| }
\hline
\multicolumn{6}{||c||}{Trends of \quotes{attention} term in arXiv} \\
\hline
Semester & Noun 1 & Noun 2 & Noun 3 & Noun 4 & Noun 5 \\
\hline
\hline
2017 S1 & saliency+context attention & content+location attention & saliency+context & saliency+ context & saliency pooling \\
 & (12.8) & (9.2) & (4.5) & (3.9) & (3.5) \\
\hline
2017 S2 & phrase hypothesis & vgg feature extractor & lstm attention mechanism & nach & example alignment \\
 & (1.5) & (1.5) & (1.3) & (1.1) & (1.0) \\
\hline
2018 S1 & gru attention & ctc attention model & propose ctc model & couple attention layer & refer article \\
 & (6.6) & (3.5) & (3.0) & (2.8) & (2.6) \\
\hline
2018 S2 & aem model & seq2seq+copying & seq2 seq+ attention & database split & compress operation \\
 & (9.6) & (4.9) & (3.2) & (2.3) & (1.8) \\
\hline
2019 S1 & w3act & w1qe & ddata & w2qe & nested attention \\
 & (4.0) & (4.0) & (2.8) & (2.0) & (1.5) \\
\hline
2019 S2 & joeynmt & mann+md & fdsm & machine translation process & japanese squ ad \\
 & (1.5) & (1.0) & (1.0) & (1.0) & (1.0) \\
\hline
2020 S1 & direct attention control & sloutsky & transformer ranker & trans+copy & travel consultation system \\
 & (1.5) & (1.5) & (1.3) & (1.3) & (1.3) \\
\hline
2020 S2 & mle+beam search & mle+sampling & standard multihead attention & lipschitz multihead attention & sense context gru \\
 & (6.3) & (4.1) & (4.0) & (3.6) & (3.3) \\
\hline
2021 S1 & govreport & hepos & hepos attention & sinkhorn encoder & decoder query \\
 & (4.3) & (2.3) & (2.0) & (2.0) & (2.0) \\
\hline
2021 S2 & attention tripnet & cagam & tripnet & triplet recognition & nwoye \\
 & (7.6) & (6.1) & (5.3) & (4.6) & (3.6) \\
\hline
\end{tabular}
\caption{Evolution of the target term \quotes{attention}. Since the first semester of 2021 we can see the emergence of the attention concept; then its spreading to the RNN-based and visual covnet architectures in the second semester of 2017 (vgg) and in the first semsest of 2018 (gru, lstm). The evolution continues to a proliferation of models, application, and finally sampling schema in 2020. Finishing with the apparition of the triplet attention.
}
\end{table}

\begin{table}[h]
\centering
\scriptsize
\begin{tabular}{ ||m{1cm}|m{2.2cm}|m{2.2cm}|m{2.2cm}|m{2.2cm}|m{2.2cm}|| }
\hline
\multicolumn{6}{||c||}{Trends of \quotes{self-attention} term in arXiv} \\
\hline
Semester & Noun 1 & Noun 2 & Noun 3 & Noun 4 & Noun 5 \\
\hline
\hline
2017 S1 & lng device & third output & initial cantilever angle & lng device behavior & cantilever angle \\
 & (3.3) & (1.8) & (1.5) & (1.3) & (1.0) \\
\hline
2017 S2 & ootp & behavioral contract & convenient normal form & fpbq & diversity enhancement \\
 & (2.1) & (1.0) & (1.0) & (1.0) & (1.0) \\
\hline
2018 S1 &  &  &  &  &  \\
 & (nan) & (nan) & (nan) & (nan) & (nan) \\
\hline
2018 S2 & attention output layer & less language & identify relevant information & k-th source sentence & side document \\
 & (1.0) & (1.0) & (1.0) & (1.0) & (0.8) \\
\hline
2019 S1 & reducibility tree & assign formula & adelfe & self-reducibility tree & polynomialtime computable function \\
 & (2.1) & (1.5) & (1.3) & (1.3) & (1.3) \\
\hline
2019 S2 & self-orthogonality & runtime goal & symbiotic relation & self-duality & mohamed bouye \\
 & (2.0) & (1.9) & (1.5) & (1.5) & (1.5) \\
\hline
2020 S1 & reproductive rate & hardmax x i & i 3 & weiwei wang & full transformer architecture \\
 & (1.8) & (1.5) & (1.5) & (1.1) & (1.0) \\
\hline
2020 S2 & self-bounding & standard probabilistic bound & self-bound function & word adapter & awareness functionality \\
 & (5.0) & (2.6) & (1.8) & (1.5) & (1.3) \\
\hline
2021 S1 & cetacean intelligence & romanians & scale dot-product attention & finetuning bert & cube flow \\
 & (4.6) & (2.0) & (1.5) & (1.0) & (1.0) \\
\hline
2021 S2 & submatrix detection & lorentz space & model expression & mw15 & average aggregation method \\
 & (1.3) & (1.0) & (1.0) & (1.0) & (0.5) \\
\hline
\end{tabular}
\caption{Evolution of the target term \quotes{self-attention}. There is too much noise in this table to see the evolution. But, we can still see some attempts to improve self-attention, such as reducibility tree, self-orthogonality and self-binding.
}
\end{table}

\begin{table}[h]
\centering
\scriptsize
\begin{tabular}{ ||m{1cm}|m{2.2cm}|m{2.2cm}|m{2.2cm}|m{2.2cm}|m{2.2cm}|| }
\hline
\multicolumn{6}{||c||}{Trends of \quotes{transformer model} term in arXiv} \\
\hline
Semester & Noun 1 & Noun 2 & Noun 3 & Noun 4 & Noun 5 \\
\hline
\hline
2017 S1 &  &  &  &  &  \\
 & (nan) & (nan) & (nan) & (nan) & (nan) \\
\hline
2017 S2 &  &  &  &  &  \\
 & (nan) & (nan) & (nan) & (nan) & (nan) \\
\hline
2018 S1 & direct translation pair & large chinese & byte pair encoding algorithm & sensitive tokenize bleu score & chinese input \\
 & (1.0) & (1.0) & (1.0) & (1.0) & (0.5) \\
\hline
2018 S2 & triangle positional embedding & target word embedding matrix & high rise & transformer nmt & only corpora \\
 & (1.8) & (1.0) & (1.0) & (1.0) & (1.0) \\
\hline
2019 S1 & rater majority & runtime infeasible & lutional neural network & traditional chinese corpus & fp16 range \\
 & (1.5) & (1.3) & (1.3) & (1.3) & (1.0) \\
\hline
2019 S2 & backward perplexity & uctx & pipeline slt & substantial undertaking & backward generation \\
 & (1.8) & (1.5) & (1.5) & (1.0) & (1.0) \\
\hline
2020 S1 & seq2seq learning problem & next sentence generation & oracle selection & jyz & metaphoric output \\
 & (1.5) & (1.5) & (1.5) & (1.5) & (1.3) \\
\hline
2020 S2 & rnn nmt & adap mt shared task icon & spm30 & odqa model & salient span masking \\
 & (1.5) & (1.3) & (1.3) & (1.0) & (1.0) \\
\hline
2021 S1 & receipt task & generate target text & align fact & noisy attribute & to bert \\
 & (2.0) & (1.0) & (1.0) & (1.0) & (1.0) \\
\hline
2021 S2 & gnn+ transformer model & gnn+ transformer & general knowledge graph & base extraction framework & equity pledge \\
 & (2.0) & (2.0) & (1.5) & (1.5) & (1.3) \\
\hline
\end{tabular}
\caption{Evolution of the target term \quotes{transformer model}. We can see from the second semester of 2018 an improvement in embedding. The evolution move to better output rating in first part of 2019. It continues with output decoding in the second part of the year, In early 2021 the focus in on universality of T5-like models, finally the evolution move to attempts to connect transformers and knowledge graphs through graph neural networks.
}
\end{table}

\begin{table}[h]
\centering
\scriptsize
\begin{tabular}{ ||m{1cm}|m{2.2cm}|m{2.2cm}|m{2.2cm}|m{2.2cm}|m{2.2cm}|| }
\hline
\multicolumn{6}{||c||}{Trends of \quotes{large language model} term in arXiv} \\
\hline
Semester & Noun 1 & Noun 2 & Noun 3 & Noun 4 & Noun 5 \\
\hline
\hline
2017 S1 &  &  &  &  &  \\
 & (nan) & (nan) & (nan) & (nan) & (nan) \\
\hline
2017 S2 & strong effective channel gain & level differential privacy guarantee & input decoding & nrf beam & federate averaging algorithm \\
 & (1.0) & (1.0) & (0.5) & (0.5) & (0.3) \\
\hline
2018 S1 &  &  &  &  &  \\
 & (nan) & (nan) & (nan) & (nan) & (nan) \\
\hline
2018 S2 &  &  &  &  &  \\
 & (nan) & (nan) & (nan) & (nan) & (nan) \\
\hline
2019 S1 & eat2seq & pre-train self & global labels & convolutional decomposition & abstract language model \\
 & (2.3) & (1.0) & (1.0) & (1.0) & (1.0) \\
\hline
2019 S2 & nra algorithm & propose mixout & deep association & purpose language representation & text continuation \\
 & (1.5) & (1.0) & (1.0) & (1.0) & (1.0) \\
\hline
2020 S1 & plausible human & gpt2 architecture & multilingual pretraine model & shot cross-lingual transferability & meaningful gap \\
 & (1.0) & (1.0) & (1.0) & (1.0) & (1.0) \\
\hline
2020 S2 & transfer m & partial game & intent detection model & hard adaptation region & shiu \\
 & (1.5) & (1.5) & (1.3) & (1.3) & (1.0) \\
\hline
2021 S1 & riddlesense & p qrnn student & metaprompt & decay instance & exist supervise system \\
 & (1.6) & (1.5) & (1.5) & (1.3) & (1.0) \\
\hline
2021 S2 & cross-lingual stance detection & transliterate dataset & local multicast model & explicit embedding space & sentence entailment \\
 & (2.0) & (1.5) & (1.5) & (1.0) & (1.0) \\
\hline
\end{tabular}
\caption{Evolution of the target term \quotes{large language model}. The LLM term properly emerge from the end of 2020; mostly for specific applications. In the following semester we see a prompting exploration emergence (meta prompt), as well as performance on complex tasks (riddlesense). Finally, the evolution focus on the development and benchmarking of multilingual models.
}
\end{table}

\begin{table}[h]
\centering
\scriptsize
\begin{tabular}{ ||m{1cm}|m{2.2cm}|m{2.2cm}|m{2.2cm}|m{2.2cm}|m{2.2cm}|| }
\hline
\multicolumn{6}{||c||}{Trends of \quotes{fine-tuning} term in arXiv} \\
\hline
Semester & Noun 1 & Noun 2 & Noun 3 & Noun 4 & Noun 5 \\
\hline
\hline
2017 S1 &  &  &  &  &  \\
 & (nan) & (nan) & (nan) & (nan) & (nan) \\
\hline
2017 S2 &  &  &  &  &  \\
 & (nan) & (nan) & (nan) & (nan) & (nan) \\
\hline
2018 S1 &  &  &  &  &  \\
 & (nan) & (nan) & (nan) & (nan) & (nan) \\
\hline
2018 S2 &  &  &  &  &  \\
 & (nan) & (nan) & (nan) & (nan) & (nan) \\
\hline
2019 S1 &  &  &  &  &  \\
 & (nan) & (nan) & (nan) & (nan) & (nan) \\
\hline
2019 S2 &  &  &  &  &  \\
 & (nan) & (nan) & (nan) & (nan) & (nan) \\
\hline
2020 S1 &  &  &  &  &  \\
 & (nan) & (nan) & (nan) & (nan) & (nan) \\
\hline
2020 S2 &  &  &  &  &  \\
 & (nan) & (nan) & (nan) & (nan) & (nan) \\
\hline
2021 S1 &  &  &  &  &  \\
 & (nan) & (nan) & (nan) & (nan) & (nan) \\
\hline
2021 S2 &  &  &  &  &  \\
 & (nan) & (nan) & (nan) & (nan) & (nan) \\
\hline
\end{tabular}
\caption{Evolution of target term \quotes{fine-tuning}. Suprinsgly the search give nothing, even though the very frequent usage of fine-tuning in the LLM domain.
}
\end{table}

\begin{table}[h]
\centering
\scriptsize
\begin{tabular}{ ||m{1cm}|m{2.2cm}|m{2.2cm}|m{2.2cm}|m{2.2cm}|m{2.2cm}|| }
\hline
\multicolumn{6}{||c||}{Trends of \quotes{transfer learning} term in arXiv} \\
\hline
Semester & Noun 1 & Noun 2 & Noun 3 & Noun 4 & Noun 5 \\
\hline
\hline
2017 S1 & original english model & reinitialize weight & overall test score & correspond letter & introspection technique \\
 & (1.0) & (1.0) & (0.5) & (0.5) & (0.3) \\
\hline
2017 S2 & multi- relevance transfer learning & common feature cluster & most exist domain adaptation method & xiaoming jin & tree covariance matrix \\
 & (3.0) & (1.0) & (1.0) & (1.0) & (1.0) \\
\hline
2018 S1 & regularize transfer learning framework & sa2017 & background depression & domain lstd & feature heatmap \\
 & (2.0) & (1.5) & (1.5) & (1.5) & (1.3) \\
\hline
2018 S2 & software engineering stack exchange & knowledge extraction task & cnn transfer learning & contrast dess & mention future work \\
 & (3.4) & (2.5) & (2.0) & (2.0) & (2.0) \\
\hline
2019 S1 & heavy channel mismatch & lilleberg & iterative transfer learning & biomedical translation task & pretrained seq sleep net \\
 & (2.0) & (1.5) & (1.0) & (1.0) & (1.0) \\
\hline
2019 S2 & birdcall & german annotate datum & sound spectrogram & project class & multilingual co ve \\
 & (2.6) & (2.0) & (1.5) & (1.5) & (1.3) \\
\hline
2020 S1 & general transfer learning & parent target language & target language scenario & child performance & arg kp \\
 & (5.1) & (4.5) & (1.8) & (1.8) & (1.5) \\
\hline
2020 S2 & largescale human & hot goal & simulate domestic environment & hot figure & point cloud noise \\
 & (2.0) & (1.6) & (1.5) & (1.5) & (1.3) \\
\hline
2021 S1 & source shapley value & mnli development & good group subset selection & target adversarial training & bi-text datum \\
 & (2.3) & (1.5) & (1.5) & (1.3) & (1.0) \\
\hline
2021 S2 & variant transfer learning & tvtl & tvtl & incoming target datum & corresponding excess risk \\
 & (4.5) & (2.8) & (2.3) & (2.0) & (2.0) \\
\hline
\end{tabular}
\caption{Evolution of the target term \quotes{transfer learning}. Even an high level of noise in this table, we can still see terms that suggest the usage of transfer learning to learn code generation capabilities in end of 2018, but it is too early according to the experts. In the end of 2019 the extracted compound nouns suggest language-to-language transfer.
}
\end{table}

\begin{table}[h]
\centering
\scriptsize
\begin{tabular}{ ||m{1cm}|m{2.2cm}|m{2.2cm}|m{2.2cm}|m{2.2cm}|m{2.2cm}|| }
\hline
\multicolumn{6}{||c||}{Trends of \quotes{conversational agent} term in arXiv} \\
\hline
Semester & Noun 1 & Noun 2 & Noun 3 & Noun 4 & Noun 5 \\
\hline
\hline
2017 S1 & institutional action & oce an & modal representation &  &  \\
 & (2.0) & (1.3) & (0.3) & (nan) & (nan) \\
\hline
2017 S2 & world user interaction & socialbot capable & major unsolved problem & milabot & emotional sentence \\
 & (2.0) & (1.0) & (1.0) & (1.0) & (1.0) \\
\hline
2018 S1 & dialog flow & alexa skills kit & world user interaction & socialbot capable & rule hand \\
 & (1.5) & (1.0) & (1.0) & (1.0) & (0.5) \\
\hline
2018 S2 & take behavior & rational framework & savitsky & axel brando & expect expense \\
 & (1.0) & (1.0) & (0.5) & (0.5) & (0.3) \\
\hline
2019 S1 & size adaption method & technological landscape & drive conversational agent & propose navigation algorithm & technical support service \\
 & (1.5) & (1.0) & (1.0) & (1.0) & (1.0) \\
\hline
2019 S2 & dialogues corpus & drive dialog model & effective conversation & engage response & erkenntnis \\
 & (1.0) & (1.0) & (1.0) & (1.0) & (1.0) \\
\hline
2020 S1 & user feedback datum & cultural barrier & kristina & semi-supervised speech recognition & skill available \\
 & (1.3) & (1.0) & (1.0) & (1.0) & (1.0) \\
\hline
2020 S2 & xspc & zts+20 & robotic hardware & perseverance rover & reference eneko agirre \\
 & (1.0) & (1.0) & (1.0) & (1.0) & (1.0) \\
\hline
2021 S1 & woe bot & generate informative response & generative teaching networks & deep learning challenge & scale conversation \\
 & (3.5) & (1.0) & (1.0) & (1.0) & (1.0) \\
\hline
2021 S2 & shot language coordination & pure cotton & sample negative response & response selection module & powerful communication model \\
 & (2.8) & (1.3) & (1.0) & (0.8) & (0.5) \\
\hline
\end{tabular}
\caption{Evolution of the target term \quotes{conversational agent}. We can observe a slow development starting from 2018, with an interest in n-shot conversion in late 2021. In overall the results are mostly noise.
}
\end{table}

\end{document}